\documentclass{article} % For LaTeX2e
\usepackage{iclr2025_conference,times}

\usepackage{amsmath,amsfonts,bm}

\def\eqref#1{equation~\ref{#1}}
\def\1{\bm{1}}

\DeclareMathAlphabet{\mathsfit}{\encodingdefault}{\sfdefault}{m}{sl}
\SetMathAlphabet{\mathsfit}{bold}{\encodingdefault}{\sfdefault}{bx}{n}

\usepackage{hyperref}
\usepackage{url}
\usepackage{enumitem}
\usepackage{graphicx}
\usepackage{bm}
\usepackage{booktabs}
\usepackage{multirow}
\usepackage{colortbl}

\usepackage{wrapfig}
\usepackage{adjustbox}
\usepackage{subcaption}

\usepackage{pifont}% http://ctan.org/pkg/pifont
\usepackage{enumitem}
\let\oldding\ding% Store old \ding in \oldding
\renewcommand{\ding}[2][1]{\scalebox{#1}{\oldding{#2}}}

\newcommand{\rebcolor}[1]{{\color{black} #1}}

\usepackage{titlesec}
\titlespacing\section{0pt}{0pt plus 0pt minus 0pt}{0pt plus 0pt minus 0pt}
\titlespacing\subsection{0pt}{0pt plus 0pt minus 0pt}{0pt plus 0pt minus 0pt}
\titlespacing\subsubsection{0pt}{0pt plus 0pt minus 0pt}{0pt plus 0pt minus 0pt}

\title{DriveTransformer: Unified Transformer for Scalable End-to-End Autonomous Driving}

\author{Xiaosong Jia*, Junqi You*, Zhiyuan Zhang*,  Junchi Yan$^{\dagger}$ \\
Sch. of Computer Science \& Sch. of Artificial Intelligence, Shanghai Jiao Tong University \\
* Equal Contributions \quad\quad $^{\dagger}$ 
 Correspondence Author \\
 \normalsize{
\color{magenta}\url{https://github.com/Thinklab-SJTU/DriveTransformer/}
    } \\ 
}

\iclrfinalcopy % Uncomment for camera-ready version, but NOT for submission.
\begin{document}

\maketitle

\begin{abstract}
End-to-end autonomous driving (E2E-AD) has emerged as a trend in the field of autonomous driving, promising a data-driven, scalable approach to system design. However, existing E2E-AD methods usually adopt the sequential paradigm of perception-prediction-planning, which leads to cumulative errors and training instability. The manual ordering of tasks also limits the system’s ability to leverage synergies between tasks (for example, planning-aware perception and game-theoretic interactive prediction and planning). Moreover, the dense BEV representation adopted by existing methods brings computational challenges for long-range perception and long-term temporal fusion.  To address these challenges, we present \textbf{DriveTransformer}, a simplified E2E-AD framework for the ease of scaling up, characterized  by three key features: \textit{Task Parallelism} (All agent, map, and planning queries direct interact with each other at each block), \textit{Sparse Representation} (Task queries direct interact with raw sensor features), and \textit{Streaming Processing} (Task queries are stored and passed as history information). As a result, the new framework is composed of three unified operations: task self-attention, sensor cross-attention, temporal cross-attention, which significantly reduces the complexity of system and leads to better training stability. \textbf{DriveTransformer} achieves state-of-the-art performance in both simulated closed-loop benchmark Bench2Drive and real world open-loop benchmark nuScenes with high FPS.
\end{abstract}

\section{Introduction}

\begin{figure}[!h]
    \centering
\includegraphics[width=1.0\linewidth]{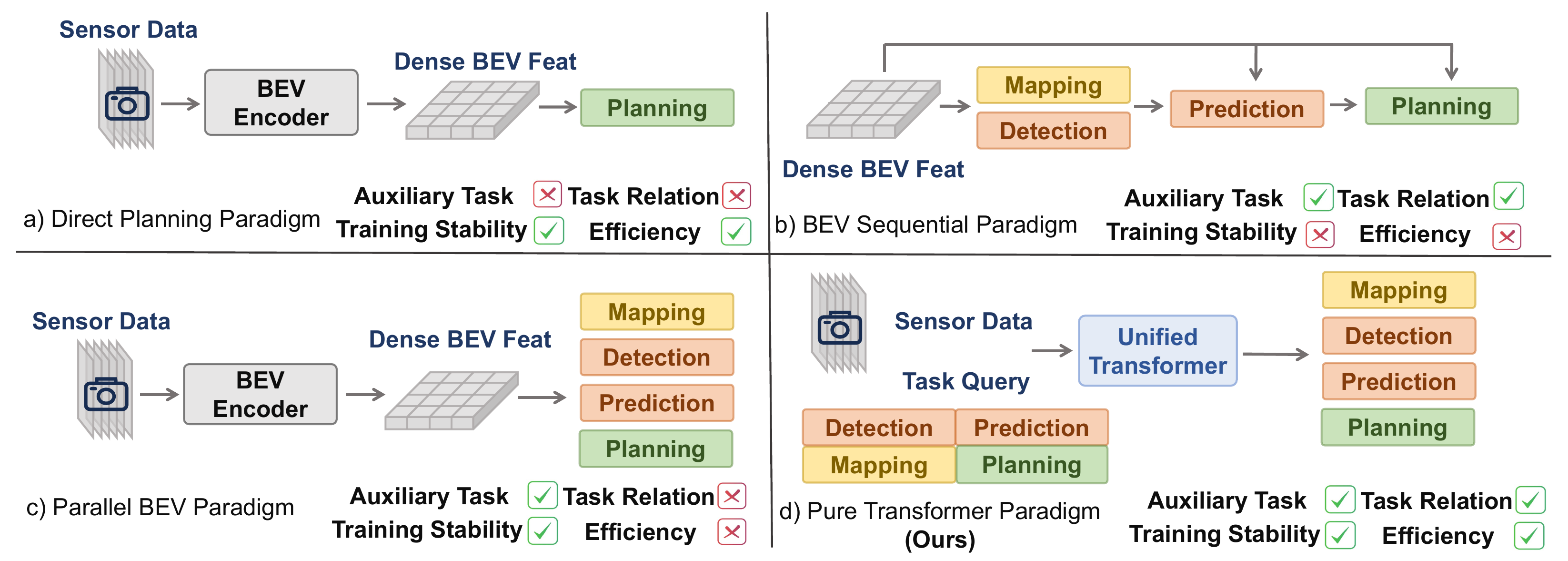}
    \caption{\textbf{End-to-End Autonomous Driving Paradigm Comparison.} The proposed pure Transformer paradigm avoids the construction of expensive BEV features and allows the tasks to learn their relations with raw sensor inputs, other tasks, and histories all by Transformer attention.}
    \label{fig:paradigm}
\end{figure}
\renewcommand{\thefootnote}{} % Temporarily remove numbering for this footnote
\footnotetext{Correspondence author is also affiliated with Shanghai lnnovation Institute. This work was in part supported by by NSFC (62222607) and Shanghai Municipal Science and Technology Major Project under Grant 2021SHZDZX0102.} % Footnote text
\renewcommand{\thefootnote}{\arabic{footnote}}

Autonomous driving has been a topic of interest~\citep{li2023delving,Yang2023LLM4DriveAS} in recent years, with significant progress being made in the field~\citep{hu2023planning,jia2023driveadapter}.  One of the most exciting approaches is end-to-end autonomous driving (E2E-AD), which aims to integrate perception~\citep{li2023delving}, prediction~\citep{jia2023towards}, and planning~\citep{li2024think2drive} into a single, holistic framework. \textbf{E2E-AD is particularly appealing due to its data-driven~\citep{lu2024activead} and scalable nature, allowing for continuous improvement with more data}.

\begin{figure}[!t]
    \centering
\includegraphics[width=1.0\linewidth]{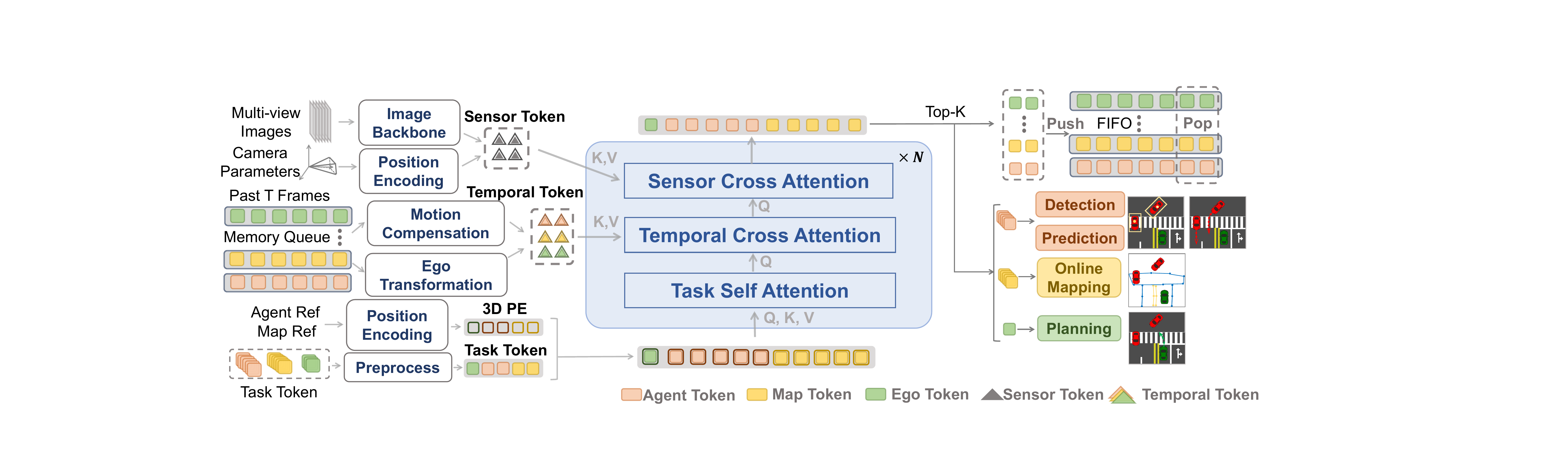}
    \caption{\textbf{Overall Framework of DriveTransformer.} DriveTransformer features streaming, parallel, and sparse token interaction. At each layer, task tokens interact with each other by \textit{Task Self Attention}, extract information from raw sensor inputs by \textit{Sensor Cross Attention}, and fuse temporal information from history task tokens in memory queue by \textit{Temporal Cross Attention}.}\vspace{-10mm}
    \label{fig:overall}
\end{figure}

Despite these advantages, existing E2E-AD methods~\citep{hu2023planning,jiang2023vad} mostly adopt a sequential pipeline of perception-prediction-planning, where downstream tasks are heavily dependent on upstream queries. \textbf{This sequential design can lead to cumulative errors and thus training instability}. For instance, the training process of UniAD~\citep{hu2023planning} necessitates a multi-stage approach: first, pre-training the BEVFormer encoder~\citep{li2022bevformer}; then, training  TrackFormer and MapFormer; and finally, training  MotionFormer and Planner. This fragmented training approach increases the complexity and difficulty of deploying and scaling the system in industrial settings. Moreover, \textbf{the manual ordering of tasks can restrict the system’s ability to leverage synergies}, such as planning-aware perception~\citep{philion2020learning,jia2023thinktwice} and game-theoretic interactive prediction and planning~\citep{ide-net,huang2023gameformer}.

Another challenge existing methods grapple with is the spatial-temporal complexity of the real world. BEV-based representations~\citep{li2022bevformer} encounter computational challenges with long-range detection~\citep{jiang2024far3d} due to the dense nature of the BEV grid.  Additionally, the image backbone of BEV methods is under-optimized due to weak gradient signals~\citep{yang2023bevformer}, hindering their ability to scale. For temporal fusion, BEV-based methods typically store history BEV features for fusion, which is computationally extensive as well~\citep{Park2022TimeWT}.  In summary, \textbf{BEV-based methods ignore the sparsity of 3D space and drop the task query of each frame}, which results in significant waste of computation and thus suffer from efficiency~\citep{Li2023IsES}.

The latest work ParaDrive~\citep{Weng2024paradrive} tries to mitigate the instability issue by removing all tasks' connection. However, it still suffers from the expensive BEV representation and their experiments is limited to open-loop, which could not reflect actual planning ability~\citep{Li2023IsES}.

To address these deficiencies, we introduce \textbf{DriveTransformer}, a framework for efficient and scalable end-to-end autonomous driving, featuring three key properties as shown in Fig.~\ref{fig:overall}:
\begin{itemize}[leftmargin=10pt, topsep=0pt, itemsep=1pt, partopsep=1pt, parsep=1pt]
\item \textbf{Task Parallelism}: All task queries directly interact  with each other at each block, fostering cross-task knowledge transfer while maintaining system stability without explicit hierarchy. 

\item \textbf{Sparse Representation}: Task queries directly engage with raw sensor features, offering an efficient and direct means of information extraction, aligning with end-to-end optimization paradigm.

\item \textbf{Streaming Processing}: Temporal fusion is achieved through a first-in-first-out queue that stores task queries in history and temporal cross attention, ensuring efficiency and feature reuse.
\end{itemize}

DriveTransformer offers \rebcolor{a unified}, parallel, and synergistic approach to E2E-AD, facilitating easier training and scalability. As a result, \rebcolor{DriveTransformer} achieves state-of-the-art closed-loop performance in Bench2Drive~\citep{jia2024bench} under CARLA simulation and state-of-the-art open-loop planning  performance on nuScenes~\citep{nuscenes} dataset.

\section{Related Works}
The concept of E2E-AD could date back to 1980s~\citep{pomerleau1988alvinn}. CIL~\citep{codevilla2018cil} trained a simple CNN to map front-view camera images directly to control commands. Refined by CILRS~\citep{codevilla2019exploring}, it incorporated an auxiliary task to predict the ego vehicle's speed, addressing issues related to inertia. PlanT~\citep{Renz2022CORL} approach suggested leveraging a Transformer architecture for the teacher model, while LBC~\citep{chen2020learning} focused on initially training a teacher model with privileged inputs. Moving forward, studies such as \cite{zhang2021roach,li2024think2drive} ventured into reinforcement learning to create driving policies. Building on these advancements,  student models were developed~\citep{wu2022trajectoryguided,hu2022model}.  In subsequent research, the use of multiple sensors became prevalent, enhancing the models' capabilities. Transfuser~\citep{Prakash2021CVPR,Chitta2022PAMI} utilized a Transformer for the integration of camera and LiDAR data. LAV~\citep{chen2022lav} adopted the PointPainting~\citep{vora2020pointpainting} technique, and Interfuser~\citep{shao2022interfuser} incorporated safety-enhanced rules into the decision-making process. Further innovations included the usage of VectorNet for map encoding by MMFN~\citep{zhang2022mmfn} and the introduction of a DETR-like scalable decoder paradigm by ThinkTwice~\citep{jia2023thinktwice} for student models. ReasonNet~\citep{shao2023reasonnet} proposed specialized modules to improve the exploitation of temporal and global information, while \cite{Jaeger2023ICCV} suggested a classification-based approach to student's output to mitigate the averaging issue.

In another branch where AD sub-tasks are explicitly conducted, ST-P3~\citep{hu2022stp3} integrated detection, prediction, and planning tasks into \rebcolor{a unified} BEV segmentation framework. Further, UniAD~\citep{hu2023planning} employed Transformer to link different tasks, and VAD~\citep{jiang2023vad} proposed a vectorized representation space. ParaDrive~\citep{Weng2024paradrive} removes the links among all tasks while BEVPlanner~\citep{Li2023IsES} removes all middle tasks.  Concurrent to our work, there are sparse query based methods~\citep{zhang2024sparseadsparsequerycentricparadigm,sun2024sparsedriveendtoendautonomousdriving,su2024difsdegocentricfullysparse}. However, they still follow the sequential pipeline while the proposed DriveTransformer unifies all tasks into parallel Transformer paradigm.

\section{Method}
Given raw sensor inputs (e.g., multi-view images), \textbf{\rebcolor{DriveTransformer}} aims to output results for multiple tasks, including object detection~\citep{Li2022DelvingIT}, motion prediction~\citep{jia2023hdgt}, online mapping~\citep{chen2022persformer}, and planning~\citep{li2024think2drive}. Each task is handled by its corresponding queries, which directly interact with each other, extract information from  raw sensor inputs, and integrate information from histories. The overall framework is illustrated in Fig.~\ref{fig:overall}.

\begin{figure}[!t]
    \centering
\includegraphics[width=1.0\linewidth]{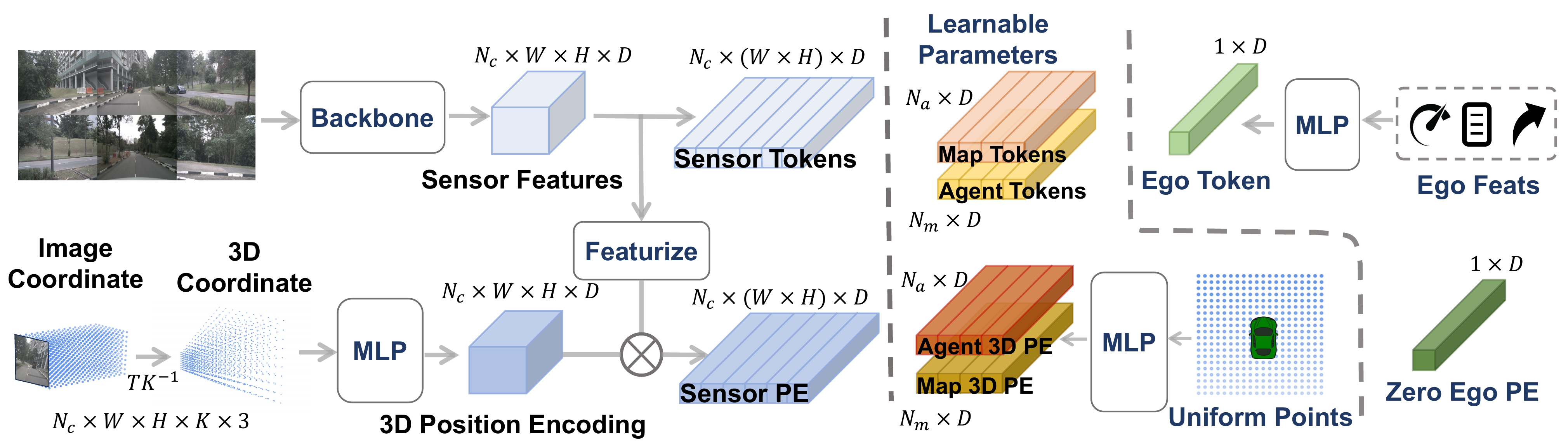}
\caption{\textbf{Initialization \& Tokenization Process.} Sensor inputs are processed by backbone while their PE are their 3D coordinate. Agent and Map tokens are initialized from learnable parameters while their initial PE are uniformly initialized. Ego token is initialized from canbus information while its PE is initialized as zeros.\vspace{-6mm}}
    \label{fig:tokenization}
\end{figure}

\subsection{Initialization \& Tokenization}
Prior to information exchange in DriveTransformer, all inputs are converted into \rebcolor{a unified} representation - tokens. Inspired by DAB-DETR~\citep{liu2022dabdetr}, all tokens consist of two parts: \textbf{semantic embeddings} for semantic information and \textbf{position encodings} for spatial localization. In Fig.~\ref{fig:tokenization}, we illustrate the process and we give details below.

\textbf{Sensor Tokens}: Multi-view images from surrounding cameras are separately encoded by backbones like ResNet~\citep{he2016deep} into  $\bm{H}_\text{sensor} \in \mathbb{R}^{N_c \times H \times W \times D}$ semantic embeddings, where $N_c$ is the number of cameras, $H$ and $W$ are the height and width of the feature map after patchification, $D$ is the hidden dimension. To encode the position information of sensor features, we adopt 3D Position Encoding in~\citep{liu2022petr,liu2022petrv2}. Specifically, for each patch with pixel coordinate (i, j), its corresponding ray in the 3D space could be represented with $K$ equally spaced 3D points: $\textbf{Ray}_{i,j} = \{TK^{-1}[i, j, d_k] | k=1,2,...,K\}$, where $T$ and $K$ are the extrinsic and intrinsic matrix of the camera, $d_k$ is the depth value for the $k^{\text{th}}$ 3D point. Then, coordinates of points in the same ray  are concatenated and fed into an MLP to obtain the position encoding for each patch~\citep{liu2022petrv2} and the 3D position encoding for all image patches is denoted as $\text{\textbf{PE}}_\text{sensor} \in \mathbb{R}^{N_c \times H \times W \times D}$

\textbf{Task Tokens}: To model the heterogeneous driving scene, three types of \textbf{task queries} are initialized to extract different information: (I) \textbf{Agent Queries} represent  dynamic objects (vehicles, pedestrians, etc), which will be used to conduct object detection and motion prediction. (II) \textbf{Map Queries} represent  static elements (lanes, traffic signs, etc), which will be used to conduct online mapping. (III) \textbf{Ego Query} represents the potential behavior of the ego vehicle, which will be used to conduct planning.  Following DAB-DETR~\citep{liu2022dabdetr}, both agent queries' and map queries' semantic embeddings are initialized randomly as learnable parameters  $\bm{H}_\text{agent} \in \mathbb{R}^{N_a \times D}$ and $\bm{H}_\text{map} \in \mathbb{R}^{N_m \times D}$ where $N_a$ and $N_m$ are the number of agent and map queries - pre-defined hyper-parameters. Their position encodings $\text{\textbf{PE}}_\text{agent} \in \mathbb{R}^{N_a \times D}$ and $\text{\textbf{PE}}_\text{map} \in \mathbb{R}^{N_m \times  D}$ are initialized uniformly within pre-defined perception range.  For planning query, its semantic embedding is initialized from an MLP encoding canbus information $\bm{H}_\text{ego} = \text{MLP}(\bm{H}_\text{canbus}) \in \mathbb{R}^{D}$ similar to BEVFormer~\citep{li2022bevformer} while its  position encoding $\text{\textbf{PE}}_\text{ego} \in \mathbb{R}^{D}$ is initialized as all zeros.

\subsection{Token Interaction}
\begin{figure}[!t]
    \centering
\includegraphics[width=1.0\linewidth]{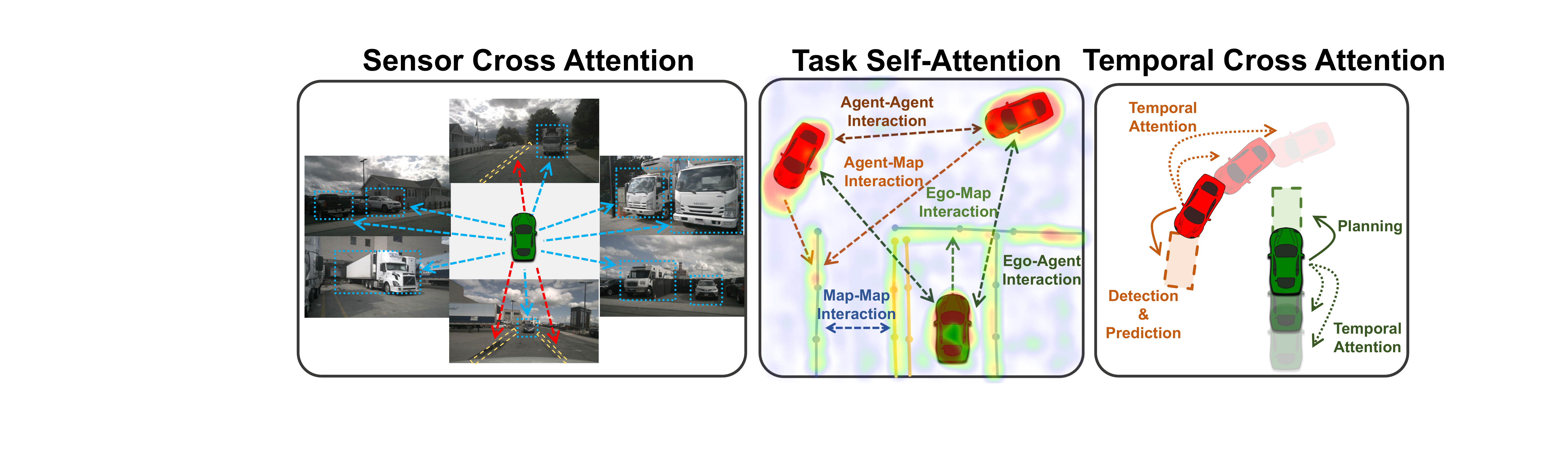}
    \caption{\textbf{Three Types of Attention in DriveTransformer.} Sensor Cross Attention provides a direct way for all tasks to access raw inputs in an end-to-end way.  Task Self-Attention allows the information exchange among tasks. Temporal Cross Attention utilizes the history states as priors\vspace{-8mm}.}
    \label{fig:attention}
\end{figure}

\textbf{All information exchange within DriveTransformer is established by the vanilla attention mechanisms~\citep{Vaswani2017AttentionIA}, ensuring scalability and easy deployment}. As a result, the model could be trained under one stage and demonstrate strong scalability, which will be shown in the experiments section. In following sub-sections, we describe the three types of information exchange adopted at each layer of DriveTransformer and the illustration is in Fig.~\ref{fig:attention}

\noindent\textbf{Sensor Cross Attention (SCA)} establishes a direct pathway between tasks and raw sensor inputs, enabling end-to-end learning without information loss. SCA is conducted as:
\begin{equation}
\setlength{\abovedisplayskip}{0pt}
\setlength{\belowdisplayskip}{0pt}
\setlength{\abovedisplayshortskip}{0pt}
\setlength{\belowdisplayshortskip}{0pt}
    \begin{split}
    \bm{H}^{\prime}_\text{ego}, \bm{H}^{\prime}_\text{agent}, \bm{H}^{\prime}_\text{map} = \text{SCA-Attention}(&Q=[\bm{H}_\text{ego} + \text{\textbf{PE}}_\text{ego}, \bm{H}_\text{agent} + \text{\textbf{PE}}_\text{agent}, \bm{H}_\text{map} + \text{\textbf{PE}}_\text{map}], \\ & K=\bm{H}_\text{sensor}+\text{\textbf{PE}}_\text{sensor}, \quad V=\bm{H}_\text{sensor})
    \end{split}
\end{equation}
where $\bm{H}^{\prime}$ denotes the updated query. In this way, raw sensor tokens are matched  by task queries based on both semantic and spatial relations to extract task-specific information in an end-to-end way without information loss. Notably, by adopting 3D position encoding~\citep{liu2022petrv2}, \textbf{DriveTransformer avoids the construction of BEV feature, which is efficient and has less gradient vanishing issue~\citep{yang2023bevformer}, allowing scaling up}.

\noindent\textbf{Task Self-Attention (TSA)} enables direct interaction among arbitrary tasks without explicit constraint, promoting synergy such as planning-aware perception~\citep{philion2020learning} and game-theoretic interactive prediction and planning~\citep{huang2023gameformer}. TSA is conducted as:
\begin{equation}
\setlength{\abovedisplayskip}{0pt}
\setlength{\belowdisplayskip}{0pt}
\setlength{\abovedisplayshortskip}{0pt}
\setlength{\belowdisplayshortskip}{0pt}
    \begin{split}
    \bm{H}^{\prime}_\text{ego}, & \bm{H}^{\prime}_\text{agent}, \bm{H}^{\prime}_\text{map}  = \text{TSA-Attention}(Q=[\bm{H}_\text{ego} + \text{\textbf{PE}}_\text{ego}, \bm{H}_\text{agent} + \text{\textbf{PE}}_\text{agent}, \bm{H}_\text{map} + \text{\textbf{PE}}_\text{map}], \\ & K=[\bm{H}_\text{ego} + \text{\textbf{PE}}_\text{ego}, \bm{H}_\text{agent} + \text{\textbf{PE}}_\text{agent}, \bm{H}_\text{map} + \text{\textbf{PE}}_\text{map}], \quad V=[\bm{H}_\text{ego}, \bm{H}_\text{agent}, \bm{H}_\text{map}])
    \end{split}
\end{equation}

By eliminating manually designed task dependencies, \textbf{the interleaved relations among tasks could be flexibly learnt via attention in a data-driven way}, which eases the scaling up. In contrast, UniAD~\citep{hu2023planning} has to adopt a multi-stage training strategy due to the inconsistency at the early stage of training where inaccurate upstream modules influence downstream modules and finally collapse the whole training.

\begin{figure}[!t]
    \centering
\includegraphics[width=1.0\linewidth]{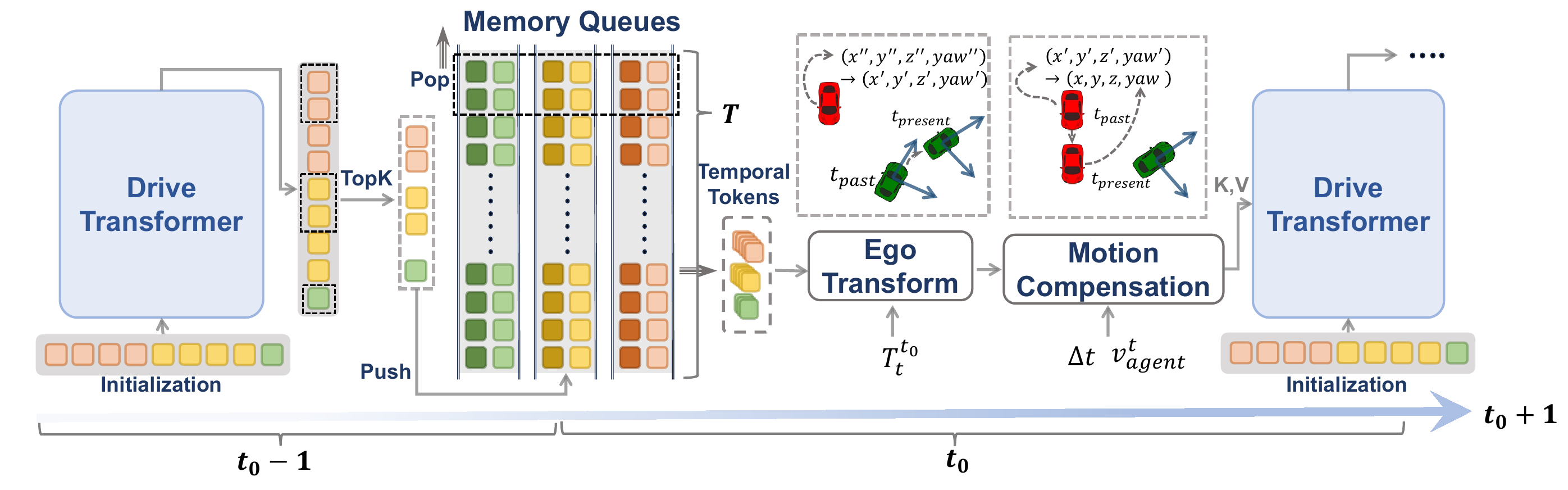}
    \caption{\textbf{Streaming Temporal Mechanism in DriveTransformer.} Top-K task queries at previous timestep from the last layer of DriveTransformer are pushed into FIFO queues. Task queries and positions are transformed into current ego coordinate system and are compensated for potential movement before feeding into Temporal Cross Attention as Key and Value. 
 \vspace{-9mm}}
    \label{fig:streaming}
\end{figure}

\noindent\textbf{Temporal Cross Attention} integrates information from previously observed history. Existing paradigms~\citep{hu2023planning,jiang2023vad} use history BEV features to pass temporal information, which introduces two drawbacks: (a) Maintaining long-term BEV features is expensive~\citep{Park2022TimeWT} (b) Previous task queries carrying strong prior semantic and spatial information are wastefully discarded. Inspired by  StreamPETR~\citep{wang2023exploring}, DriveTransformer maintains First-In-First-Out (FIFO) queues of queries for each task respectively and conducts cross attention to history queries in the queue at each layer to fuse temporal information, as illustrated in Fig.~\ref{fig:streaming}.

Specifically, denote $\bm{H}_\text{ego}^t$, $\bm{H}_\text{agent}^t$, $\bm{H}_\text{map}^t$ with their corresponding position encodings $\textbf{PE}_\text{ego}^t$,  $\text{\textbf{PE}}_\text{agent}^t$, $\text{\textbf{PE}}_\text{map}^t$ as the ego queries, agent queries, and map queries of the final layer of DriveTransformer at time-step $t$. Suppose current time-step is $t_0$ and we maintain FIFO queues $\text{Queue}_{\text{ego}} = \{(\bm{H}_\text{ego}^t, \textbf{PE}_\text{ego}^t) | t=t_0-1, t_0-2, ..., t_0-T_\text{queue}\}$,  $\text{Queue}_{\text{agent}} = \{(\bm{H}_\text{agent}^t, \textbf{PE}_\text{agent}^t) | t=t_0-1, t_0-2, ..., t_0-T_\text{queue}\}$ and $\text{Queue}_{\text{map}} = \{(\bm{H}_\text{map}^t, \textbf{PE}_\text{map}^t) | t=t_0-1, t_0-2, ..., t_0-T_\text{queue}\}$ where $T_\text{queue}$ is a pre-set hyper-parameter to control the length of temporal queue. After each time-step, current task queries at the final layer are pushed into the queue and the task queries at $t_0-T$ are popped out. Further, since there are redundant queries in DETR style methods~\citep{carion2020endtoend}, for agent and map queries, only those with top-K confidence scores are kept, where K is a hyper-parameter.

Temporal Cross Attention uses the history queries as Key and Value. Since the ego reference point at different time-step could be different, the history queries' PE is transformed into current coordinate system (Ego Transformation):
\begin{equation}
\setlength{\abovedisplayskip}{0pt}
\setlength{\belowdisplayskip}{0pt}
\setlength{\abovedisplayshortskip}{0pt}
\setlength{\belowdisplayshortskip}{0pt}
    \hat{\text{\textbf{PE}}}^t= \text{MLP}(T_t^{t_0}\text{\textbf{Pos}}^t) \quad \text{where} \; t=t_0-1, t_0-2, ..., t_0-T_\text{queue}
\end{equation}

where $\hat{\text{\textbf{PE}}^t}$ is the transformed PE, $T_t^{t_0}$ is the coordinate transformation matrix from $t$ to $t_0$. Besides, since other agents could have their own movement, following~\citep{wang2023exploring}, we conduct DiT~\citep{Peebles2022DiT} style ada-LN for Motion Compensation:
\begin{equation}
\setlength{\abovedisplayskip}{4pt}
\setlength{\belowdisplayskip}{4pt}
\setlength{\abovedisplayshortskip}{4pt}
\setlength{\belowdisplayshortskip}{4pt}
\hat{\text{\textbf{PE}}}^t_\text{agent} = \text{LayerNorm}(\hat{\text{\textbf{PE}}^t_{\text{agent}}}, [\gamma,\beta]=\text{MLP}(v^t_\text{agent} * (t-t_0))) \; \text{where} \; t=t_0-1,..., t_0-T_\text{queue}
\end{equation}
where the layer-norm's weight $\gamma$ and bias $beta$ is controlled by the predicted velocity of agents at time-step $t$ and the time interval between $t$ and current time-step $t_0$. Besides, we also set the relative time embedding as $t_{\text{emb}} = \text{MLP}(t-t_0)$ to indicate different time-steps and 
Temporal Cross-Attention is conducted as:
\begin{equation}
\setlength{\abovedisplayskip}{0pt}
\setlength{\belowdisplayskip}{0pt}
\setlength{\abovedisplayshortskip}{0pt}
\setlength{\belowdisplayshortskip}{0pt}
    \begin{split}
    \bm{H}^{\prime}_\text{task}  & = \text{TCA-Attention}(Q=\bm{H}_\text{task} + \text{\textbf{PE}}_\text{task},\;  K=\{\bm{H}^t_\text{task}+\hat{\text{\textbf{PE}}}^t_\text{task}+t_\text{emb}|t=t_0-1,..., t_0-T_\text{queue}\}, \\ & V=\{\bm{H}^t_\text{task}|t=t_0-1,..., t_0-T_\text{queue}\}) \; \text{where task=Ego, Map, Agent}
    \end{split}
\end{equation}

\textbf{Pure Attention Architecture: In summary, DriveTransformer is a stack of multiple blocks where each block contains the aforementioned three attentions and a FFN}:

\begin{equation}
    \begin{split}
    \bm{H}^{l+1}_\text{ego}, \bm{H}^{l+1}_\text{agent}, \bm{H}^{l+1}_\text{map} = \text{FFN}(\text{TSA}((\text{TCA}(\text{SCA}&([\bm{H}^{l}_\text{ego}, \bm{H}^{l}_\text{agent}, \bm{H}^{l}_\text{map}], \bm{H}_\text{sensor}), \bm{H}^{\prime}_\text{task}), \\ &[\text{Queue}_{\text{ego}}, \text{Queue}_{\text{agent}}, \text{Queue}_{\text{map}}]))
    \end{split}
\end{equation}
where $l$ and $l+1$ is the layer index,  FFN means MLP in Transformer~\citep{Vaswani2017AttentionIA} and we omit the PE, residual connection, and pre-layernorm for brevity. Note that the raw sensor tokens $\bm{H}_\text{sensor}$ and history information $\text{Queue}_{\text{ego}}, \text{Queue}_{\text{agent}}, \text{Queue}_{\text{map}}$ are shared across all blocks.

\begin{figure}[!t]
    \centering
\includegraphics[width=1.0\linewidth]{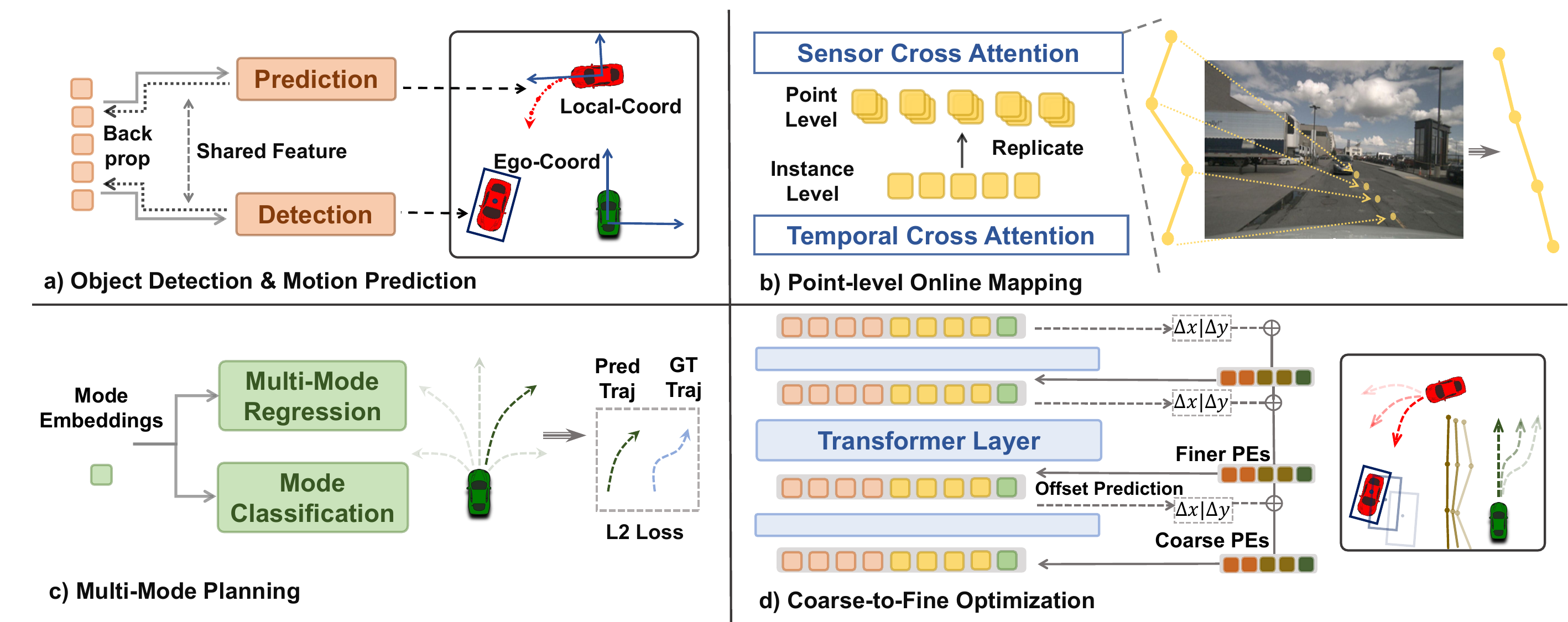}
\caption{\textbf{Task Head Designs.} (a) Detection and Motion share the same agent feature, which associate objects without tracking. Trajectory prediction in local coordinate system further disentangles the two tasks, ensuring training stability. (b) Map points in the same polyline have different PEs in Sensor Cross Attention to retrieve fine-grained features. (c) Planning head combines the ego query with different mode embeddings to conduct multi-mode prediction. (d) Tasks' positions are gradually refined and their corresponding PEs are gradually more accurate, providing better interaction.\vspace{-5mm}}
    \label{fig:taskhead}
\end{figure}

\subsection{DETR-Style Task Head}
Inspired by DETR~\citep{carion2020endtoend,liu2022dabdetr}, \textbf{task heads are set  after each block to gradually refine the predictions and the PE would be correspondingly updated}. In  following sub-sections, we introduce task specific designs and the updating strategy of PE, as shown in Fig.~\ref{fig:taskhead}.

\noindent\textbf{Object Detection \& Motion Prediction}: Existing E2E methods~\citep{zeng2022motr,hu2023planning} still adopt the classic detection-association-prediction pipeline, which introduces instability to training due to the inherent difficulty of association~\citep{weng2022whose}. For example, in UniAD, there must be a 3D object detection pretrained BEVFormer to avoid the divergence of TrackFormer and then the MapFormer and TrackFormer must be primarily trained before the end-to-end training of MotionFormer and planning head, which necessitates its multi-stage training strategy and thus hinders the scaling up.  

To alleviate this issue, DriveTransformer adopts a more end-to-end methodology: \textbf{conducting object detection and motion prediction without tracking, by feeding the same agent query into different task heads}. The same feature for the same agent naturally establish  associations between detection and prediction.  For temporal association, since Temporal Cross Attention is conducted between current tokens and \emph{all} history tokens, the explicit association is avoided, replaced by the learning-based attention mechanism. To further improve the training stability and reduce the interference between these two tasks, the label of motion prediction is transformed to the local coordinate system of each agent~\citep{jia2022multi} and thus its loss is not influenced by detection results at all.  Only during inference, the waypoints from the prediction are transformed into the global coordinate system based on the detection to calculate motion prediction related metrics.

\noindent\textbf{Online Mapping} Recent progress in the sub-field~\citep{li2023lanesegnet,liu2024leveraging} suggests the importance of point-level instead of instance-level feature retrieval due to the irregular and diverse distribution of map polylines. Thus, when conducting \textit{Sensor Cross Attention}, we replicate each map query for $N_\text{point}$ times paired with position encodings for each single point. In this way, for those long polylines, \textbf{each point could retrieve raw sensor information with better locality}. To integrate  separate point-level map queries into instance-level ones for other modules, we adopt a light-weight PointNet~\citep{qi2017pointnet} with max-pooling and MLPs.

\noindent\textbf{Planning}: We model the the future movement of the ego vehicle as Gaussian Mixture Model to avoid mode averaging, as widely done in the motion prediction field~\citep{liang2020learning}. Specifically, we divide all training trajectories according to  their direction and distance into six categories: Go Straight, Stop, Left Turn, Sharp Left Turn, Right Turn, Sharp Right Turn. To  generate trajectories of these modes, six mode embeddings are generated by feeding their sine\&cosine encoded position~\citep{Vaswani2017AttentionIA} into an MLP and then we add them to the ego feature to predict six mode-specific ego trajectories. During training, only the Ground-Truth mode's trajectory would be used to calculate regression loss, i.e. winner-take-all~\citep{liang2020learning}, and we also train a classification head to predict current mode. During inference, the trajectory from the mode with the highest confidence score would be used to compute metrics or execute.% \jxs{May not do that} Moreover, as predictions of other agents are also in the local coordinate system, we share the mode query and head of two tasks - prediction and ego planning, which aligns with~\citep{jia2022multi,chen2022lav}. \jxs{TODO:visualization}

\noindent\textbf{Coarse-to-Fine Optimization}: The success of DETR series~\citep{carion2020endtoend} demonstrates the power of end-to-end learning by coarse-to-fine optimization. In DriveTransformer, the Position Encodings (PE) of all task queries are updated after each block based on current predictions. Specifically, \textbf{the Map and Agent PE are encoded by their corresponding predicted positions and semantic classes with an MLP to capture the spatial and semantic relations among elements}. The ego PE is encoded by \textbf{the predicted planning trajectory with an MLP to capture the ego intentions for the possible interactions}. Similar to DETR, During training, losses are applied on task heads at all blocks while during inference we only use the output from the last block.

\subsection{Loss \& Optimization}
\textbf{DriveTransformer is trained under one single stage}, where each tasks could gradually learn to find out their relations in Task Self-Attention, without collapsing each others' basic convergence  under Sensor Cross Attention and Temporal Cross Attention. There are detection loss (DETR-style hungarian matching loss~\citep{carion2020endtoend}), prediction loss (winner-take-all style loss~\citep{liang2020learning}), online mapping loss (MapTR~\citep{MapTR} style hungarian matching loss), and planning loss (winner-take-all style loss) where we adjust weights to make sure all terms have the same magnitude around 1, as in the following equation:
\begin{equation}
   \mathcal{L_\text{overall}} = w_\text{detection}\mathcal{L_\text{detection}} + w_\text{motion}\mathcal{L_\text{motion}} + w_\text{mapping}\mathcal{L_\text{mapping}} + w_\text{planning}\mathcal{L_\text{planning}}
\end{equation}

\begin{table}[tb!]
\centering
\small
\caption{\textbf{Planning Performance in Bench2Drive}. Avg. L2 is averaged over the predictions in 2 seconds under 2Hz. * denotes expert feature distillation. All latency are measured by the averaged inference step-time on CARLA evaluation in A6000. \vspace{-2mm} \label{tab:b2d}}
 \resizebox{\linewidth}{!}{
\begin{tabular}{l|c|cccc|c}
\toprule
\multirow{2}{*}{\textbf{Method}} & \textbf{Open-loop Metric} & \multicolumn{4}{c|}{\textbf{Closed-loop Metric}} & \multirow{2}{*}{\textbf{Latency}} \\ \cmidrule{2-6} 
                                 &                  Avg. L2 $\downarrow$           & \cellcolor{gray!30} Driving Score $\uparrow$  & Success Rate(\%) $\uparrow$ & Efficiency $\uparrow$ & Comfortness $\uparrow$\\ \midrule
AD-MLP~\citep{zhai2023ADMLP}                           & 3.64              & \cellcolor{gray!30} 18.05     &  0.00  & 48.45 &   22.63 & \textbf{3ms}  \\ 
UniAD-Tiny~\citep{hu2023planning}                               &  0.80       & \cellcolor{gray!30} 40.73    & 13.18 & 123.92 & 47.04 & 420.4ms \\
UniAD-Base~\citep{hu2023planning}                            &  0.73          & \cellcolor{gray!30} 45.81     & 16.36 & 129.21 & 43.58 & 663.4ms  \\

VAD~\citep{jiang2023vad}                              &            0.91                 & \cellcolor{gray!30} 42.35     & 15.00 & \textbf{157.94} & \textbf{46.01} & 278.3ms \\
DriveTransformer-Large (\textbf{Ours}) & \textbf{0.62} & \cellcolor{gray!30} \textbf{63.46} &  \textbf{35.01} & 100.64 & 20.78 & 211.7ms \\ \midrule

TCP*~\citep{wu2022trajectoryguided}    & 1.70                & \cellcolor{gray!30}40.70     & 15.00  & 54.26 & 47.80 & \textbf{83ms} \\ 
TCP-ctrl*                              & -                 &  \cellcolor{gray!30}30.47    & 7.27  & 55.97 & \textbf{51.51} & 83ms \\
TCP-traj*    & 1.70                &  \cellcolor{gray!30} 59.90     & 30.00 & 76.54 & 18.08  & 83ms  \\
TCP-traj w/o distillation                              & 1.96                &  \cellcolor{gray!30} 49.30     & 20.45  & \textbf{78.78} & 22.96 & 83ms \\
ThinkTwice*~\citep{jia2023thinktwice}                             & \textbf{0.95}               &  \cellcolor{gray!30} 62.44     & 31.23  & 69.33 & 16.22 & 762ms  \\
DriveAdapter*~\citep{jia2023driveadapter}                            & 1.01                &  \cellcolor{gray!30}\textbf{64.22}     & \textbf{33.08} & 70.22 & 16.01 & 931ms   \\ \bottomrule

\end{tabular}}
\vspace{-4mm}
\end{table}

\begin{table}[tb!]
\centering
\caption{\textbf{Multi-Ability Results of E2E-AD Methods.}  * denotes expert feature distillation.\label{tab:ability}\vspace{-2mm}}
\resizebox{\linewidth}{!}{
\begin{tabular}{l|ccccc|c}
\toprule
\multirow{2}{*}{\textbf{Method}} & \multicolumn{5}{c}{\textbf{Ability} (\%) $\uparrow$}                                                                                                                \\ \cmidrule{2-7} 
                                 & \multicolumn{1}{c}{Merging} & \multicolumn{1}{c}{Overtaking} & \multicolumn{1}{c}{Emergency Brake} & \multicolumn{1}{c}{Give Way} & Traffic Sign & \textbf{Mean} \\ \midrule
AD-MLP~\citep{zhai2023ADMLP}        & 0.00        & 0.00            & 0.00        & 0.00           &  0.00    & \cellcolor{gray!30}0.00         \\
UniAD-Tiny~\citep{hu2023planning}   & 7.04        & 10.00           & 21.82       &  20.00         & 14.61    & \cellcolor{gray!30}14.69        \\ 
UniAD-Base~\citep{hu2023planning}       & 12.16           & 20.00       &  23.64         & 10.00    & 13.89 & \cellcolor{gray!30}15.94       \\ 
VAD~\citep{jiang2023vad}            & 7.14        & 20.00          & 16.36       &  20.00         & 20.22    & \cellcolor{gray!30}16.75       \\
DriveTransformer-Large (\textbf{Ours}) & 17.57 & 35.00 &  48.36 & 40.00 &  52.10 & \cellcolor{gray!30}\textbf{38.60} \\  \midrule
TCP*~\citep{wu2022trajectoryguided}        & 17.50        & 13.63           & 20.00        &  10.00         & 6.81     & \cellcolor{gray!30}13.59        \\
TCP-ctrl*                           & 9.23        & 5.00           & 9.10        &  10.00          & 6.81     & \cellcolor{gray!30}8.03         \\
TCP-traj*                           & 12.50       & 22.73           & 52.72       &  40.00      & 46.63    & \cellcolor{gray!30}34.92        \\ 
TCP-traj w/o distillation                           & 14.71       & 7.50           & 38.18      &  50.00        & 29.97   & \cellcolor{gray!30}28.03        \\ 
ThinkTwice*~\citep{jia2023thinktwice}                           & 13.72       &  22.93          & 52.99       &  50.00        & 47.78    & \cellcolor{gray!30}37.48        \\ 
DriveAdapter*~\citep{jia2023driveadapter}                          & 14.55       & 22.61          & 54.04      &  50.00         & 50.45   & \cellcolor{gray!30}\textbf{38.33}        \\   \bottomrule
\end{tabular}}\vspace{-4mm}
\end{table}

\begin{table}[]
\begin{center}
\caption{\textbf{Open-loop planning performance in nuScenes under VAD metric}. $^\dagger$ denotes LiDAR-based methods. $^\ddagger$ denotes the usage of ego-status. \label{tab:nuscenes}}
\vspace{-3mm}
% \resizebox{0.9\textwidth}{!}{
\begin{tabular}{l|cccc|cccc}
\toprule
\multirow{2}{*}{Method} &
\multicolumn{4}{c|}{L2 (m) $\downarrow$} & 
\multicolumn{4}{c}{Collision (\%) $\downarrow$} \\
& 1s & 2s & 3s & Avg. & 1s & 2s & 3s & Avg. \\
\midrule
NMP~\citep{zeng2019nmp}$\dagger$ & - & - & 2.31 & \cellcolor{gray!30}- & - & - & 1.92 & \cellcolor{gray!30}-  \\
SA-NMP~\citep{zeng2019nmp}$\dagger$ & - & - & 2.05 & \cellcolor{gray!30}- & - & - & 1.59 & \cellcolor{gray!30}- \\
FF~\citep{hu2021ff}$\dagger$ & 0.55 & 1.20 & 2.54 & \cellcolor{gray!30}1.43 & 0.06 & 0.17 & 1.07 & \cellcolor{gray!30}0.43  \\
EO~\citep{khurana2022eo}$\dagger$ & 0.67 & 1.36 & 2.78 & \cellcolor{gray!30}1.60 & 0.04 & 0.09 & 0.88 & \cellcolor{gray!30}0.33 \\
\midrule
ST-P3~\citep{hu2022stp3} & 1.33 & 2.11 & 2.90 & \cellcolor{gray!30}2.11 & 0.23 & 0.62 & 1.27 & \cellcolor{gray!30}0.71  \\
UniAD~\citep{hu2023planning} & 0.48 & 0.74 & 1.07 & \cellcolor{gray!30}0.76 & 0.12 & 0.13 & 0.28 & \cellcolor{gray!30}0.17  \\
VAD-Tiny~\citep{jiang2023vad} & 0.46 & 0.76 & 1.12 & \cellcolor{gray!30}0.78 & 0.21 & 0.35 & 0.58 & \cellcolor{gray!30}0.38  \\
VAD-Base~\citep{jiang2023vad} & 0.41 & 0.70 & 1.05 & \cellcolor{gray!30}0.72 & 0.07 & 0.17 & 0.41 & 
\cellcolor{gray!30}0.22 \\
BEVPlaner~\citep{Li2023IsES} & 0.27 & 0.54 & 0.90 & \cellcolor{gray!30}0.57 & - & - & - & 
\cellcolor{gray!30}- \\
\rebcolor{DriveTransformer-Large (\textbf{Ours})} & \rebcolor{0.19} & \rebcolor{0.34} & \rebcolor{0.66} & \cellcolor{gray!30}\rebcolor{0.40} & \rebcolor{0.03} & \rebcolor{0.10} & \rebcolor{0.21} & 
\cellcolor{gray!30}\rebcolor{0.11} \\
\midrule
VAD-Tiny$^\ddagger$~\citep{jiang2023vad} & 0.20 & 0.38 & 0.65 & \cellcolor{gray!30}0.41 & 0.10 & 0.12 & 0.27 & \cellcolor{gray!30}0.16  \\
VAD-Base$^\ddagger$~\citep{jiang2023vad} & 0.17 & 0.34 & 0.60 & \cellcolor{gray!30}0.37 & 0.07 & 0.10 & 0.24 & \cellcolor{gray!30}0.14 \\
AD-MLP$^\ddagger$~\citep{zhai2023ADMLP} & 0.20 & \textbf{0.26} & \textbf{0.41} & \cellcolor{gray!30}\textbf{0.29} & 0.17 & 0.18 & 0.24 & \cellcolor{gray!30}0.19 \\

BEVPlaner++$^\ddagger$~\citep{Li2023IsES} & \textbf{0.16} & 0.32 & 0.57 & \cellcolor{gray!30}0.35 & - & - & - & \cellcolor{gray!30}- \\
ParaDrive$^\ddagger$~\citep{Weng2024paradrive} & 0.25 & 0.46 & 0.74 & \cellcolor{gray!30}0.48 & 0.14 & 0.23 & 0.39 & \cellcolor{gray!30}0.25 \\
DriveTransformer-Large$^\ddagger$ (\textbf{Ours}) & \textbf{0.16} & 0.30 & 0.55 & \cellcolor{gray!30}0.33 & \textbf{0.01} & \textbf{0.06} & \textbf{0.15} & \cellcolor{gray!30}\textbf{0.07}  \\ 
\bottomrule
\end{tabular}% }
\end{center}
\vspace{-2mm}
\end{table}

\section{Experiments}

\subsection{Dataset \& Benchmark} We use Bench2Drive~\citep{jia2024bench}, a closed-loop evaluation protocol under CARLA Leaderboard 2.0 for end-to-end autonomous driving. It provides an official training set, where we use the base set (1000 clips) for fair comparison with all the other baselines. We use the official 220 routes for evaluation. Additionally, we compare our method with other state-of-the-art baselines on nuScenes~\citep{caesar2020nuscenesmultimodaldatasetautonomous} open-loop evaluation. There are three different size of models:
\begin{table}[!h]
\centering
\vspace{-3mm}
\caption{\textbf{Size Configuration of DriveTransformer.} Latency is measured by the averaged model time for closed-loop evaluation in CARLA. Training batch size is measured by A800 (80G) to fill the GPU memory. Driving Score is under Dev10 benchmark.\vspace{-3mm}}
\begin{tabular}{l|c|cccc}
\toprule
Size                   & Configuration         & \#Parameters & Latency & Training Batch Size & Driving Score \\ \midrule
Small & 3 Layers, 256 Hidden  & 47.41M       & 93.8ms  & 48                  & 45.04         \\
Base  & 6 Layers, 512 Hidden  & 178.05M      & 139.6ms & 28                  & 60.45         \\
Large & 12 Layers, 768 Hidden & 646.33M      & 221.6ms & 12                  & 68.22         \\ \bottomrule
\end{tabular}
\vspace{-6mm}
\end{table}

When comparing with SOTA works, we report results of DriveTransformer-Large. For ablation studies, since evaluating on 220 routes of Bench2Drive could take days, we select 10 representative scenes (namely \textbf{Dev10}) balancing behaviors weathers, and towns and report results on it with DriveTransformer-base for quick validation.   %Implementation details are in Appendix~\ref{sec:detail} and details about Dev10 are in Appendix~\ref{sec:dev10}.

\subsection{Comparison with State-of-the-Art Methods}
We compare DriveTransformer with SOTA E2E-AD methods in Table~\ref{tab:b2d}, Table~\ref{tab:ability}, and Table~\ref{tab:nuscenes}. We  observe that DriveTransformer persistently outperforms SOTA methods. From Table~\ref{tab:b2d}, \textbf{DriveTransformer has a lower inference latency compared to UniAD and VAD}. Notably, because of the unified, sparse, and streaming Transformer design, DriveTransformer could be trained with batch size 12 in A800 (80G) while UniAD with batch size 1 and VAD with batch size 4.

\subsection{Ablation Studies}
In ablation studies, all closed-loop experiments are conducted on \textbf{Dev10}, a subset of Bench2Drive 220 routes, and all open-loop results are on Bench2Drive official validation set (50 clips). Please check Appendix~\ref{sec:dev10} for details. We use a smaller model (6 layers and 512 hidden dimension) for ablation studies to save computational resource if not specified. We will open source \textbf{Dev10} protocol, model code, and model checkpoints.

\begin{figure}[!t]
    \centering
\includegraphics[width=1.0\linewidth]{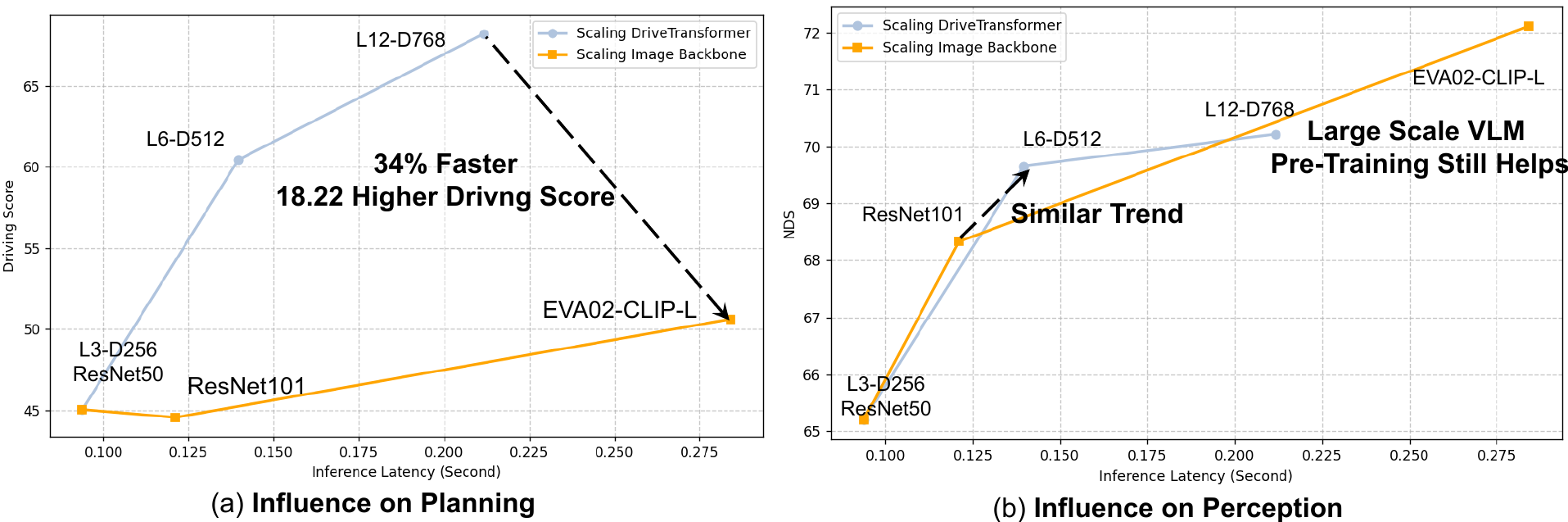}
\caption{\textbf{Scaling Study of Driving Transformer in Dev10 benchmark.} Directly scaling up the unified Transformer structure benefits most on the planning while adopting powerful image backbones, especially large-scale pretrained ones brings gains for perception.\vspace{-2mm}}
    \label{fig:scaling}
\end{figure}

\noindent\textbf{Scaling Study}: One attractive characteristic of Transformer-based paradigm is its extremely strong scalability~\citep{radford2019language,brown2020languagemodelsfewshotlearners}. Since DriveTransformer is composed of Transformers, we study the scaling behavior of increasing layers and hidden dimension simultaneously and compare the strategy with scaling image backbones similar to~\citep{hu2023planning,jiang2023vad} as shown in Fig.~\ref{fig:scaling}. We could observe that scaling up the decoder part, i.e., the number of layers and width for the three attention brings more gains compared to scale up image backbones. It is natural since the former one directly adds more computation to the planning task.  On the other hand, for the perception task, scaling up the decoder has similar trend with scaling up image backbones, which demonstrates the generalizing ability of the proposed DriveTransformer. However, it still falls behind the large scale vision-language pretraining image backbone - EVA02-CLIP-L~\citep{eva02}, aligning with findings in~\citep{wang2023exploring}. It could be an important direction to study how to combine VLLM with autonomous driving~\citep{Yang2023LLM4DriveAS}.

\begin{table}[!tb]
\centering
\caption{\textbf{Paradigm Design Study with DriveTransformer-Base in Dev10.}\vspace{-3mm}\label{tab:paradigm}} 
\subcaptionbox{\textbf{Attention Mechanism\vspace{-1mm}}\label{tab:attention}}{
\begin{adjustbox}{width=0.41\linewidth}
\begin{tabular}{l|cc}
\toprule
Method                         & \textbf{Driving Score} $\uparrow$  & \textbf{Success Rate} $\uparrow$   \\ \midrule
\cellcolor{gray!30}Full Attention    & \cellcolor{gray!30}\textbf{60.45}       & \cellcolor{gray!30}\textbf{30.00}  \\ \midrule
w/o Sensor-CA  & 8.41       & 0.00 \\
w/o Task-SA & 52.37       & 20.00  \\
w/o Temporal-CA                       & 56.22       & 20.00  \\ \bottomrule
\end{tabular}
\end{adjustbox}
}
\subcaptionbox{\textbf{Training Strategies.\vspace{-1mm}}\label{tab:training}}{
\begin{adjustbox}{width=0.45\linewidth}
\begin{tabular}{l|cc}
\toprule
Method                         & \textbf{Driving Score} $\uparrow$  & \textbf{Success Rate} $\uparrow$     \\ \midrule
\cellcolor{gray!30}DriveTransformer    & \cellcolor{gray!30}\textbf{60.45}       & \cellcolor{gray!30}\textbf{30.00} \\ \midrule
Planning Only & 54.22       & 20.00 \\
Pretrain Perception  & \textbf{60.22}       & \textbf{30.00} \\
w/o Middle Supervision  &  51.67     &  10.00   \\ \bottomrule
\end{tabular}
\end{adjustbox}
}
\vspace{-4mm}
\end{table}

\begin{table}[!t]
\centering
\caption{\textbf{Task Head Design Study with DriveTransformer-base.}\vspace{-3mm}\label{tab:taskhead}} 
\subcaptionbox{\textbf{Detection \& Prediction\vspace{-1mm}}\label{tab:prediction}}{
\begin{adjustbox}{width=0.28\linewidth}
\begin{tabular}{l|cc}
\toprule
Method                         & \textbf{minADE} $\downarrow$   \\ \midrule
\cellcolor{gray!30}Local Prediction    & \cellcolor{gray!30}\textbf{1.34}       \\
Global Prediction & 2.68 \\ \bottomrule
\end{tabular}
\end{adjustbox}
}
\subcaptionbox{\textbf{ Mapping.\vspace{-1mm}}\label{tab:mapping}}{
\begin{adjustbox}{width=0.16\linewidth}
\begin{tabular}{l|cc}
\toprule
Method                         & \textbf{mAP} $\uparrow$    \\ \midrule
\cellcolor{gray!30} Point PE    & \cellcolor{gray!30}\textbf{20.25}     \\ \midrule
 Line PE &  14.55        \\ \bottomrule
\end{tabular}
\end{adjustbox}
}
\subcaptionbox{\textbf{Planning.\vspace{-1mm}}\label{tab:planning}}{
\begin{adjustbox}{width=0.45\linewidth}
\begin{tabular}{l|cc}
\toprule
Method                         & \textbf{Driving Score} $\uparrow$  & \textbf{Success Rate} $\uparrow$     \\ \midrule
\cellcolor{gray!30}Multiple Mode    & \cellcolor{gray!30}\textbf{60.45}       & \cellcolor{gray!30}\textbf{30.00} \\ 
Single Mode & 49.19       & 20.00 \\ \bottomrule
\end{tabular}
\end{adjustbox}
}
\vspace{-5mm}
\end{table}

\noindent\textbf{Paradigm Design Study}: In Table~\ref{tab:paradigm}, we ablate the design of DriveTransformer.  We conclude that: \raisebox{-0.5pt}{\ding[1.1]{182\relax}} Based on Table~\ref{tab:attention}, \textbf{all three types of attention are helpful}. \raisebox{-0.5pt}{\ding[1.1]{183\relax}} It makes sense that Sensor Cross Attention is especially important since model would drive blindly without sensor information. \raisebox{-0.5pt}{\ding[1.1]{184\relax}}Temporal information has the least influence, which aligns with the findings in~\citep{Chitta2022PAMI}. \raisebox{-0.5pt}{\ding[1.1]{185\relax}}Task Self-Attention could improve the driving score since the ego  query could utilize the detected objects and map elements to conduct planning. \raisebox{-0.5pt}{\ding[1.1]{186\relax}} Based on Table~\ref{tab:training}, we could find that discarding  auxiliary tasks leads to performance decay, which may come from the fact that it is rather difficult to fit the single planning output with high-dimensional inputs (surrounding camera images). Actually, it is a common practice in the end-to-end autonomous driving community to adopt auxiliary tasks to regularize the learned representations~\citep{chen2022lav,Prakash2021CVPR,jia2023thinktwice}. \raisebox{-0.5pt}{\ding[1.1]{187\relax}} \textbf{One stage training is enough} for convergence and the perception pretraining, i.e, two stage training, does not provide advantages. It comes from the design that there is no manual dependency among tasks and thus all tasks could first learn from the raw sensor inputs and history information instead of influencing each others' convergence. \raisebox{-0.5pt}{\ding[1.1]{188\relax}} As a complex Transformer based framework, we find that the removal of supervision for middle layers collapse the training. It might need to further explore how to  scale up the structure with only final supervisions.

\noindent\textbf{Task Design Study}: In Table~\ref{tab:taskhead}, we ablate the designs of task heads and conclude that:
\raisebox{-0.5pt}{\ding[1.1]{182\relax}} For Table~\ref{tab:prediction}, Formulating the prediction output in the local coordinate system decouples the object detection and motion prediction. As a result, the two tasks could optimize their objectives separately while the shared input agent features naturally associate the same agent. The superior performance demonstrates the effectiveness to avoid directly predict in the global coordinate system, aligns with~\citep{jia2023hdgt,shi2024mtrmultiagentmotionprediction}.  \raisebox{-0.5pt}{\ding[1.1]{183\relax}} For Table~\ref{tab:mapping}, Point level PE in Sensor Cross Attention leads to significantly better online mapping results, which shows the effectiveness of extending the potential perception range for lane detection~\citep{liu2024leveraging,li2023lanesegnet}. \raisebox{-0.5pt}{\ding[1.1]{184\relax}} For Table~\ref{tab:planning}, multi-mode planning outperforms single mode planning explicitly. By visualization, We observe that multi-mode planning achieves better control especially on scenarios requiring subtle steers. It comes from the fact that single mode prediction with L2 loss models the output space as one single Gaussian distribution and thus suffers from mode averaging. %In contrast, multi-mode planning models the output space as Gaussian Mixture Model (GMM), which is able to generate diverse trajectories~\citep{chai2019multipath}.

\begin{figure}[!t]
    \centering
\includegraphics[width=0.8\linewidth]{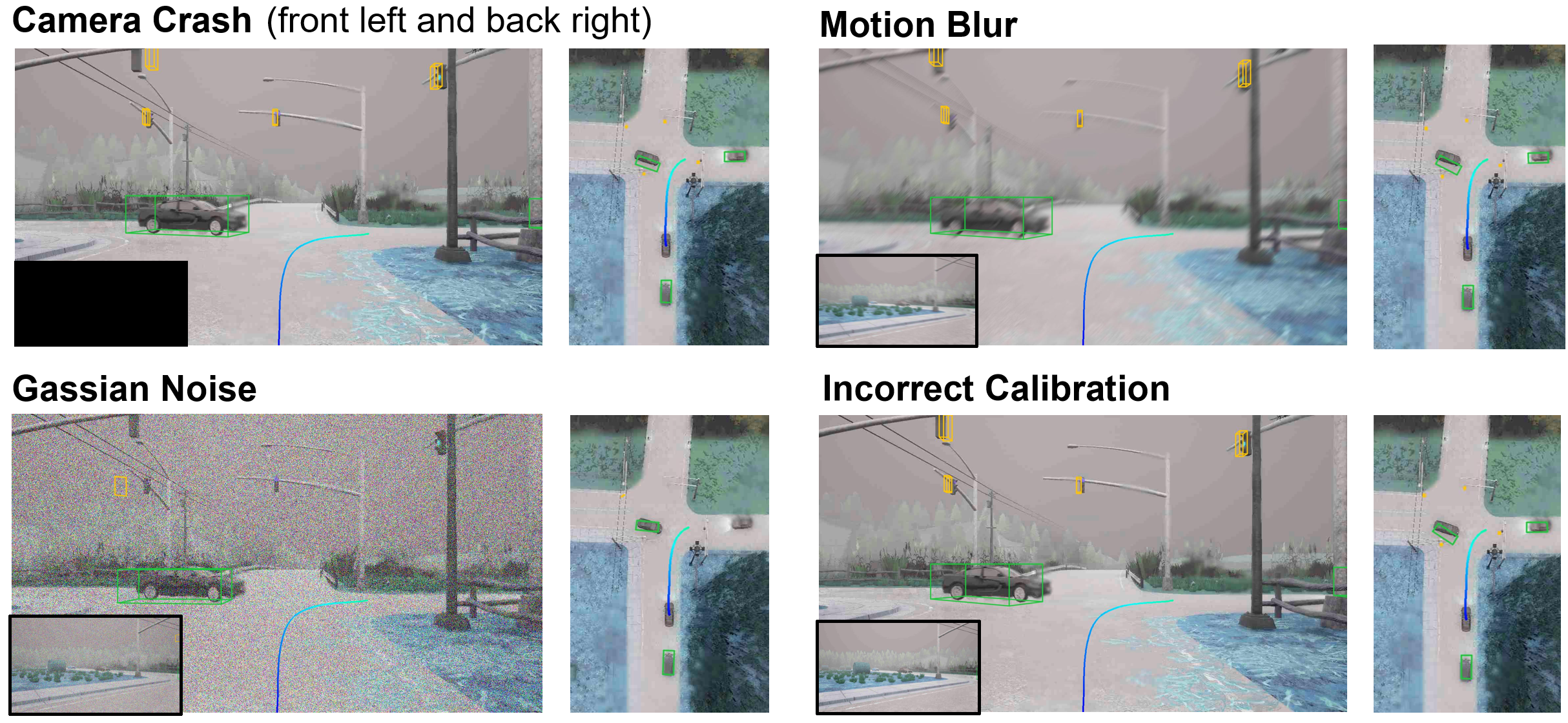}
\caption{\textbf{Visualization of Detection and Planning under Different Robust Challenges.}\vspace{-4mm}}
    \label{fig:robustness}
\end{figure}

\begin{table}[!t]
\begin{center}
\caption{\textbf{Robustness on Planning (Closed-Loop Evaluation)\label{tab:robust-planning}}. Driving Score$\uparrow$ is reported.\vspace{-2mm}} 
\resizebox{1.0\textwidth}{!}{
\begin{tabular}{l|c|cccc}
\toprule
Methods & \textbf{Regular} & \textbf{Camera Crash} & \textbf{Incorrect Calibration} & \textbf{Motion Blur} & \textbf{Gaussian Noise} \\
\midrule
VAD-Base~\citep{jiang2023vad} & 53.45 & 48.54 (9.2\%$\downarrow$ ) & 38.46 (28.04\%$\downarrow$) & 45.47 (14.93\%$\downarrow$) & 44.53 (16.72\%$\downarrow$)\\ 
DriveTransformer (Ours) & 60.45 & 58.67 (\textbf{2.9}\%$\downarrow$) & 56.53 (\textbf{5.94}\%$\downarrow$) & 54.04 (\textbf{10.60}\%$\downarrow$) & 56.94 (\textbf{6.02}\%$\downarrow$)\\  
\bottomrule
\end{tabular} }
\end{center}
\vspace{-4mm}
\end{table}

\begin{table}[!t]
\begin{center}
\caption{\textbf{Robustness on Perception (Open-Loop Evaluation)}. Object detection NDS$\uparrow$  is reported.\label{tab:robust-perception}\vspace{-2mm}} 
\resizebox{1.0\textwidth}{!}{
\begin{tabular}{l|c|cccc}
\toprule
Methods & \textbf{Regular} & \textbf{Camera Crash} & \textbf{Incorrect Calibration} & \textbf{Motion Blur} & \textbf{Gaussian Noise} \\
\midrule
VAD-Base~\citep{jiang2023vad} & 53.21 & 39.31(26.12\%$\downarrow$) & 43.01 (21.05\%$\downarrow$)  & 42.34 (22.31\%$\downarrow$) &  45.16 (17.01\%$\downarrow$)            \\
DriveTransformer (Ours) & 69.65 & 61.84 (\textbf{11.2}\%$\downarrow$) & 66.06(\textbf{5.15}\%$\downarrow$) & 61.61(\textbf{11.54}\%$\downarrow$) & 63.20(\textbf{9.26}\%$\downarrow$)\\ 
\bottomrule
\end{tabular} }
\end{center}
\vspace{-8mm}
\end{table}

\subsection{Robustness Analysis}
Autonomous driving, as an outdoor task, would frequently encounter many kinds of 
events and failures and thus it is an important perspective to examine the robustness of the system.  To this end, We adopt 4 settings in~\citep{xie2023robobev}. \raisebox{-0.5pt}{\ding[1.1]{182\relax}} Camera Crash. Two cameras are masked as all black. \raisebox{-0.5pt}{\ding[1.1]{183\relax}} Incorrect Calibration. rotation and transition noises are added to camera extrinsic parameters. \raisebox{-0.5pt}{\ding[1.1]{184\relax}} Motion Blur is applied on images.  \raisebox{-0.5pt}{\ding[1.1]{185\relax}} Gaussian Noise is applied on images. From Table~\ref{tab:robust-planning} and Table~\ref{tab:robust-perception}, DriveTransformer demonstrates significantly better robustness compared to VAD. It might be because VAD requires the construction of BEV feature, which is sensitive to perception inputs. On the other hand, \textbf{DriveTransformer directly interacts with raw sensor features and thus be able to ignore those failure or noisy inputs and demonstrates better robustness}.

\section{Conclusion}
In this work, we present DriveTransformer, a unified Transformer based paradigm for end-to-end autonomous driving, featured by task parallel, streaming processing, and sparse representation. It achieves state-of-the-art performance on both Bench2Drive in CARLA closed-loop evaluation and  nuScenes open-loop evaluation with high FPS, demonstrating the efficiency of those designs.

\newpage
\bibliography{iclr2025_conference}

\begin{thebibliography}{69}
\providecommand{\natexlab}[1]{#1}
\providecommand{\url}[1]{\texttt{#1}}
\expandafter\ifx\csname urlstyle\endcsname\relax
  \providecommand{\doi}[1]{doi: #1}\else
  \providecommand{\doi}{doi: \begingroup \urlstyle{rm}\Url}\fi

\bibitem[Brown et~al.(2020)Brown, Mann, Ryder, Subbiah, Kaplan, Dhariwal, Neelakantan, Shyam, Sastry, Askell, Agarwal, Herbert-Voss, Krueger, Henighan, Child, Ramesh, Ziegler, Wu, Winter, Hesse, Chen, Sigler, Litwin, Gray, Chess, Clark, Berner, McCandlish, Radford, Sutskever, and Amodei]{brown2020languagemodelsfewshotlearners}
Tom~B. Brown, Benjamin Mann, Nick Ryder, Melanie Subbiah, Jared Kaplan, Prafulla Dhariwal, Arvind Neelakantan, Pranav Shyam, Girish Sastry, Amanda Askell, Sandhini Agarwal, Ariel Herbert-Voss, Gretchen Krueger, Tom Henighan, Rewon Child, Aditya Ramesh, Daniel~M. Ziegler, Jeffrey Wu, Clemens Winter, Christopher Hesse, Mark Chen, Eric Sigler, Mateusz Litwin, Scott Gray, Benjamin Chess, Jack Clark, Christopher Berner, Sam McCandlish, Alec Radford, Ilya Sutskever, and Dario Amodei.
\newblock Language models are few-shot learners, 2020.
\newblock URL \url{https://arxiv.org/abs/2005.14165}.

\bibitem[Caesar et~al.(2020{\natexlab{a}})Caesar, Bankiti, Lang, Vora, Liong, Xu, Krishnan, Pan, Baldan, and Beijbom]{caesar2020nuscenesmultimodaldatasetautonomous}
Holger Caesar, Varun Bankiti, Alex~H. Lang, Sourabh Vora, Venice~Erin Liong, Qiang Xu, Anush Krishnan, Yu~Pan, Giancarlo Baldan, and Oscar Beijbom.
\newblock nuscenes: A multimodal dataset for autonomous driving, 2020{\natexlab{a}}.
\newblock URL \url{https://arxiv.org/abs/1903.11027}.

\bibitem[Caesar et~al.(2020{\natexlab{b}})Caesar, Bankiti, Lang, Vora, Liong, Xu, Krishnan, Pan, Baldan, and Beijbom]{nuscenes}
Holger Caesar, Varun Bankiti, Alex~H. Lang, Sourabh Vora, Venice~Erin Liong, Qiang Xu, Anush Krishnan, Yu~Pan, Giancarlo Baldan, and Oscar Beijbom.
\newblock nuscenes: A multimodal dataset for autonomous driving.
\newblock In \emph{CVPR}, 2020{\natexlab{b}}.

\bibitem[Carion et~al.(2020)Carion, Massa, Synnaeve, Usunier, Kirillov, and Zagoruyko]{carion2020endtoend}
Nicolas Carion, Francisco Massa, Gabriel Synnaeve, Nicolas Usunier, Alexander Kirillov, and Sergey Zagoruyko.
\newblock End-to-end object detection with transformers, 2020.

\bibitem[Chen \& Kr{\"a}henb{\"u}hl(2022)Chen and Kr{\"a}henb{\"u}hl]{chen2022lav}
Dian Chen and Philipp Kr{\"a}henb{\"u}hl.
\newblock Learning from all vehicles.
\newblock In \emph{CVPR}, 2022.

\bibitem[Chen et~al.(2020)Chen, Zhou, Koltun, and Kr{\"a}henb{\"u}hl]{chen2020learning}
Dian Chen, Brady Zhou, Vladlen Koltun, and Philipp Kr{\"a}henb{\"u}hl.
\newblock Learning by cheating.
\newblock In \emph{CoRL}, pp.\  66--75. PMLR, 2020.

\bibitem[Chen et~al.(2022)Chen, Sima, Li, Zheng, Xu, Geng, Li, He, Shi, Qiao, and Yan]{chen2022persformer}
Li~Chen, Chonghao Sima, Yang Li, Zehan Zheng, Jiajie Xu, Xiangwei Geng, Hongyang Li, Conghui He, Jianping Shi, Yu~Qiao, and Junchi Yan.
\newblock Persformer: 3d lane detection via perspective transformer and the openlane benchmark.
\newblock In \emph{European Conference on Computer Vision (ECCV)}, 2022.

\bibitem[Chitta et~al.(2022)Chitta, Prakash, Jaeger, Yu, Renz, and Geiger]{Chitta2022PAMI}
Kashyap Chitta, Aditya Prakash, Bernhard Jaeger, Zehao Yu, Katrin Renz, and Andreas Geiger.
\newblock Transfuser: imitation with transformer-based sensor fusion for autonomous driving.
\newblock \emph{TPAMI}, 2022.

\bibitem[Codevilla et~al.(2018)Codevilla, M{\"u}ller, L{\'o}pez, Koltun, and Dosovitskiy]{codevilla2018cil}
Felipe Codevilla, Matthias M{\"u}ller, Antonio L{\'o}pez, Vladlen Koltun, and Alexey Dosovitskiy.
\newblock End-to-end driving via conditional imitation learning.
\newblock In \emph{ICRA}, pp.\  4693--4700, 2018.

\bibitem[Codevilla et~al.(2019)Codevilla, Santana, L{\'o}pez, and Gaidon]{codevilla2019exploring}
Felipe Codevilla, Eder Santana, Antonio~M L{\'o}pez, and Adrien Gaidon.
\newblock Exploring the limitations of behavior cloning for autonomous driving.
\newblock In \emph{CVPR}, pp.\  9329--9338, 2019.

\bibitem[Fang et~al.(2024)Fang, Sun, Wang, Huang, Wang, and Cao]{eva02}
Yuxin Fang, Quan Sun, Xinggang Wang, Tiejun Huang, Xinlong Wang, and Yue Cao.
\newblock Eva-02: A visual representation for neon genesis.
\newblock \emph{Image and Vision Computing}, pp.\  105171, 2024.

\bibitem[He et~al.(2016)He, Zhang, Ren, and Sun]{he2016deep}
Kaiming He, Xiangyu Zhang, Shaoqing Ren, and Jian Sun.
\newblock Deep residual learning for image recognition.
\newblock In \emph{Proceedings of the IEEE conference on computer vision and pattern recognition}, pp.\  770--778, 2016.

\bibitem[Hu et~al.(2022{\natexlab{a}})Hu, Corrado, Griffiths, Murez, Gurau, Yeo, Kendall, Cipolla, and Shotton]{hu2022model}
Anthony Hu, Gianluca Corrado, Nicolas Griffiths, Zak Murez, Corina Gurau, Hudson Yeo, Alex Kendall, Roberto Cipolla, and Jamie Shotton.
\newblock Model-based imitation learning for urban driving.
\newblock \emph{NeurIPS}, 2022{\natexlab{a}}.

\bibitem[Hu et~al.(2021)Hu, Huang, Dolan, Held, and Ramanan]{hu2021ff}
Peiyun Hu, Aaron Huang, John Dolan, David Held, and Deva Ramanan.
\newblock Safe local motion planning with self-supervised freespace forecasting.
\newblock In \emph{CVPR}, 2021.

\bibitem[Hu et~al.(2022{\natexlab{b}})Hu, Chen, Wu, Li, Yan, and Tao]{hu2022stp3}
Shengchao Hu, Li~Chen, Penghao Wu, Hongyang Li, Junchi Yan, and Dacheng Tao.
\newblock St-p3: End-to-end vision-based autonomous driving via spatial-temporal feature learning.
\newblock In \emph{ECCV}, 2022{\natexlab{b}}.

\bibitem[Hu et~al.(2023)Hu, Yang, Chen, Li, Sima, Zhu, Chai, Du, Lin, Wang, et~al.]{hu2023planning}
Yihan Hu, Jiazhi Yang, Li~Chen, Keyu Li, Chonghao Sima, Xizhou Zhu, Siqi Chai, Senyao Du, Tianwei Lin, Wenhai Wang, et~al.
\newblock Planning-oriented autonomous driving.
\newblock In \emph{CVPR}, pp.\  17853--17862, 2023.

\bibitem[Huang et~al.(2023)Huang, Liu, and Lv]{huang2023gameformer}
Zhiyu Huang, Haochen Liu, and Chen Lv.
\newblock Gameformer: Game-theoretic modeling and learning of transformer-based interactive prediction and planning for autonomous driving.
\newblock In \emph{Proceedings of the IEEE/CVF International Conference on Computer Vision}, pp.\  3903--3913, 2023.

\bibitem[Jaeger et~al.(2023)Jaeger, Chitta, and Geiger]{Jaeger2023ICCV}
Bernhard Jaeger, Kashyap Chitta, and Andreas Geiger.
\newblock Hidden biases of end-to-end driving models.
\newblock In \emph{ICCV}, 2023.

\bibitem[Jia et~al.(2021)Jia, Sun, Tomizuka, and Zhan]{ide-net}
Xiaosong Jia, Liting Sun, Masayoshi Tomizuka, and Wei Zhan.
\newblock Ide-net: Interactive driving event and pattern extraction from human data.
\newblock \emph{IEEE RA-L}, 6\penalty0 (2):\penalty0 3065--3072, 2021.
\newblock \doi{10.1109/LRA.2021.3062309}.

\bibitem[Jia et~al.(2022)Jia, Sun, Zhao, Tomizuka, and Zhan]{jia2022multi}
Xiaosong Jia, Liting Sun, Hang Zhao, Masayoshi Tomizuka, and Wei Zhan.
\newblock Multi-agent trajectory prediction by combining egocentric and allocentric views.
\newblock In \emph{CoRL}, pp.\  1434--1443. PMLR, 2022.

\bibitem[Jia et~al.(2023{\natexlab{a}})Jia, Chen, Wu, Zeng, Yan, Li, and Qiao]{jia2023towards}
Xiaosong Jia, Li~Chen, Penghao Wu, Jia Zeng, Junchi Yan, Hongyang Li, and Yu~Qiao.
\newblock Towards capturing the temporal dynamics for trajectory prediction: a coarse-to-fine approach.
\newblock In \emph{CoRL}, pp.\  910--920. PMLR, 2023{\natexlab{a}}.

\bibitem[Jia et~al.(2023{\natexlab{b}})Jia, Gao, Chen, Yan, Liu, and Li]{jia2023driveadapter}
Xiaosong Jia, Yulu Gao, Li~Chen, Junchi Yan, Patrick~Langechuan Liu, and Hongyang Li.
\newblock Driveadapter: Breaking the coupling barrier of perception and planning in end-to-end autonomous driving.
\newblock In \emph{ICCV}, 2023{\natexlab{b}}.

\bibitem[Jia et~al.(2023{\natexlab{c}})Jia, Wu, Chen, Liu, Li, and Yan]{jia2023hdgt}
Xiaosong Jia, Penghao Wu, Li~Chen, Yu~Liu, Hongyang Li, and Junchi Yan.
\newblock Hdgt: Heterogeneous driving graph transformer for multi-agent trajectory prediction via scene encoding.
\newblock \emph{TPAMI}, 2023{\natexlab{c}}.

\bibitem[Jia et~al.(2023{\natexlab{d}})Jia, Wu, Chen, Xie, He, Yan, and Li]{jia2023thinktwice}
Xiaosong Jia, Penghao Wu, Li~Chen, Jiangwei Xie, Conghui He, Junchi Yan, and Hongyang Li.
\newblock Think twice before driving: Towards scalable decoders for end-to-end autonomous driving.
\newblock In \emph{CVPR}, 2023{\natexlab{d}}.

\bibitem[Jia et~al.(2024)Jia, Yang, Li, Zhang, and Yan]{jia2024bench}
Xiaosong Jia, Zhenjie Yang, Qifeng Li, Zhiyuan Zhang, and Junchi Yan.
\newblock Bench2drive: Towards multi-ability benchmarking of closed-loop end-to-end autonomous driving.
\newblock In \emph{NeurIPS 2024 Datasets and Benchmarks Track}, 2024.

\bibitem[Jiang et~al.(2023)Jiang, Chen, Xu, Liao, Chen, Zhou, Zhang, Liu, Huang, and Wang]{jiang2023vad}
Bo~Jiang, Shaoyu Chen, Qing Xu, Bencheng Liao, Jiajie Chen, Helong Zhou, Qian Zhang, Wenyu Liu, Chang Huang, and Xinggang Wang.
\newblock Vad: Vectorized scene representation for efficient autonomous driving.
\newblock \emph{ICCV}, 2023.

\bibitem[Jiang et~al.(2024)Jiang, Li, Liu, Wang, Jia, Wang, Han, and Zhang]{jiang2024far3d}
Xiaohui Jiang, Shuailin Li, Yingfei Liu, Shihao Wang, Fan Jia, Tiancai Wang, Lijin Han, and Xiangyu Zhang.
\newblock Far3d: Expanding the horizon for surround-view 3d object detection.
\newblock In \emph{Proceedings of the AAAI Conference on Artificial Intelligence}, volume~38, pp.\  2561--2569, 2024.

\bibitem[Khurana et~al.(2022)Khurana, Hu, Dave, Ziglar, Held, and Ramanan]{khurana2022eo}
Tarasha Khurana, Peiyun Hu, Achal Dave, Jason Ziglar, David Held, and Deva Ramanan.
\newblock Differentiable raycasting for self-supervised occupancy forecasting.
\newblock In \emph{ECCV}, 2022.

\bibitem[Li et~al.(2022{\natexlab{a}})Li, Sima, Dai, Wang, Lu, Wang, Xie, Li, Deng, Tian, Zhu, Chen, Li, Gao, Geng, Zeng, Li, Yang, Jia, Yu, Qiao, Lin, Liu, Yan, Shi, and Luo]{Li2022DelvingIT}
Hongyang Li, Chonghao Sima, Jifeng Dai, Wenhai Wang, Lewei Lu, Huijie Wang, Enze Xie, Zhiqi Li, Hanming Deng, Haonan Tian, Xizhou Zhu, Li~Chen, Tianyu Li, Yulu Gao, Xiangwei Geng, Jianqiang Zeng, Yang Li, Jiazhi Yang, Xiaosong Jia, Bo~Yu, Y.~Qiao, Dahua Lin, Siqian Liu, Junchi Yan, Jianping Shi, and Ping Luo.
\newblock Delving into the devils of bird’s-eye-view perception: A review, evaluation and recipe.
\newblock \emph{IEEE Transactions on Pattern Analysis and Machine Intelligence}, 46:\penalty0 2151--2170, 2022{\natexlab{a}}.

\bibitem[Li et~al.(2023)Li, Sima, Dai, Wang, Lu, Wang, Zeng, Li, Yang, Deng, et~al.]{li2023delving}
Hongyang Li, Chonghao Sima, Jifeng Dai, Wenhai Wang, Lewei Lu, Huijie Wang, Jia Zeng, Zhiqi Li, Jiazhi Yang, Hanming Deng, et~al.
\newblock Delving into the devils of bird's-eye-view perception: A review, evaluation and recipe.
\newblock \emph{IEEE Transactions on Pattern Analysis and Machine Intelligence}, 2023.

\bibitem[Li et~al.(2024{\natexlab{a}})Li, Jia, Wang, and Yan]{li2024think2drive}
Qifeng Li, Xiaosong Jia, Shaobo Wang, and Junchi Yan.
\newblock Think2drive: Efficient reinforcement learning by thinking in latent world model for quasi-realistic autonomous driving (in carla-v2).
\newblock \emph{arXiv preprint arXiv:2402.16720}, 2024{\natexlab{a}}.

\bibitem[Li et~al.(2024{\natexlab{b}})Li, Jia, Wang, Chen, Jiang, Yan, and Li]{li2023lanesegnet}
Tianyu Li, Peijin Jia, Bangjun Wang, Li~Chen, Kun Jiang, Junchi Yan, and Hongyang Li.
\newblock Lanesegnet: Map learning with lane segment perception for autonomous driving.
\newblock In \emph{ICLR}, 2024{\natexlab{b}}.

\bibitem[Li et~al.(2022{\natexlab{b}})Li, Wang, Li, Xie, Sima, Lu, Qiao, and Dai]{li2022bevformer}
Zhiqi Li, Wenhai Wang, Hongyang Li, Enze Xie, Chonghao Sima, Tong Lu, Yu~Qiao, and Jifeng Dai.
\newblock Bevformer: Learning bird’s-eye-view representation from multi-camera images via spatiotemporal transformers.
\newblock \emph{ECCV}, 2022{\natexlab{b}}.

\bibitem[Li et~al.(2024{\natexlab{c}})Li, Yu, Lan, Li, Kautz, Lu, and {\'A}lvarez]{Li2023IsES}
Zhiqi Li, Zhiding Yu, Shiyi Lan, Jiahan Li, Jan Kautz, Tong Lu, and Jos{\'e}~M. {\'A}lvarez.
\newblock Is ego status all you need for open-loop end-to-end autonomous driving?
\newblock \emph{IEEE/CVF Conference on Computer Vision and Pattern Recognition (CVPR)}, pp.\  14864--14873, 2024{\natexlab{c}}.
\newblock URL \url{https://api.semanticscholar.org/CorpusID:265664457}.

\bibitem[Liang et~al.(2020)Liang, Yang, Hu, Chen, Liao, Feng, and Urtasun]{liang2020learning}
Ming Liang, Bin Yang, Rui Hu, Yun Chen, Renjie Liao, Song Feng, and Raquel Urtasun.
\newblock Learning lane graph representations for motion forecasting.
\newblock In \emph{ECCV}, 2020.

\bibitem[Liao et~al.(2023)Liao, Chen, Wang, Cheng, Zhang, Liu, and Huang]{MapTR}
Bencheng Liao, Shaoyu Chen, Xinggang Wang, Tianheng Cheng, Qian Zhang, Wenyu Liu, and Chang Huang.
\newblock Maptr: Structured modeling and learning for online vectorized hd map construction.
\newblock In \emph{International Conference on Learning Representations}, 2023.

\bibitem[Liu et~al.(2022{\natexlab{a}})Liu, Li, Zhang, Yang, Qi, Su, Zhu, and Zhang]{liu2022dabdetr}
Shilong Liu, Feng Li, Hao Zhang, Xiao Yang, Xianbiao Qi, Hang Su, Jun Zhu, and Lei Zhang.
\newblock {DAB}-{DETR}: Dynamic anchor boxes are better queries for {DETR}.
\newblock In \emph{International Conference on Learning Representations}, 2022{\natexlab{a}}.
\newblock URL \url{https://openreview.net/forum?id=oMI9PjOb9Jl}.

\bibitem[Liu et~al.(2022{\natexlab{b}})Liu, Wang, Zhang, and Sun]{liu2022petr}
Yingfei Liu, Tiancai Wang, Xiangyu Zhang, and Jian Sun.
\newblock Petr: Position embedding transformation for multi-view 3d object detection.
\newblock \emph{arXiv preprint arXiv:2203.05625}, 2022{\natexlab{b}}.

\bibitem[Liu et~al.(2022{\natexlab{c}})Liu, Yan, Jia, Li, Gao, Wang, Zhang, and Sun]{liu2022petrv2}
Yingfei Liu, Junjie Yan, Fan Jia, Shuailin Li, Qi~Gao, Tiancai Wang, Xiangyu Zhang, and Jian Sun.
\newblock Petrv2: A unified framework for 3d perception from multi-camera images.
\newblock \emph{arXiv preprint arXiv:2206.01256}, 2022{\natexlab{c}}.

\bibitem[Liu et~al.(2024)Liu, Zhang, Liu, Zhao, and Xu]{liu2024leveraging}
Zihao Liu, Xiaoyu Zhang, Guangwei Liu, Ji~Zhao, and Ningyi Xu.
\newblock Leveraging enhanced queries of point sets for vectorized map construction.
\newblock \emph{arXiv preprint arXiv:2402.17430}, 2024.

\bibitem[Lu et~al.(2024)Lu, Jia, Xie, Liao, Yang, and Yan]{lu2024activead}
Han Lu, Xiaosong Jia, Yichen Xie, Wenlong Liao, Xiaokang Yang, and Junchi Yan.
\newblock Activead: Planning-oriented active learning for end-to-end autonomous driving, 2024.

\bibitem[Park et~al.(2023)Park, Xu, Yang, Keutzer, Kitani, Tomizuka, and Zhan]{Park2022TimeWT}
Jinhyung Park, Chenfeng Xu, Shijia Yang, Kurt Keutzer, Kris Kitani, Masayoshi Tomizuka, and Wei Zhan.
\newblock Time will tell: New outlooks and a baseline for temporal multi-view 3d object detection.
\newblock 2023.

\bibitem[Peebles \& Xie(2022)Peebles and Xie]{Peebles2022DiT}
William Peebles and Saining Xie.
\newblock Scalable diffusion models with transformers.
\newblock \emph{arXiv preprint arXiv:2212.09748}, 2022.

\bibitem[Philion et~al.(2020)Philion, Kar, and Fidler]{philion2020learning}
Jonah Philion, Amlan Kar, and Sanja Fidler.
\newblock Learning to evaluate perception models using planner-centric metrics.
\newblock In \emph{Proceedings of the IEEE/CVF Conference on Computer Vision and Pattern Recognition}, pp.\  14055--14064, 2020.

\bibitem[Pomerleau(1988)]{pomerleau1988alvinn}
Dean~A Pomerleau.
\newblock Alvinn: An autonomous land vehicle in a neural network.
\newblock \emph{NeurIPS}, 1, 1988.

\bibitem[Prakash et~al.(2021)Prakash, Chitta, and Geiger]{Prakash2021CVPR}
Aditya Prakash, Kashyap Chitta, and Andreas Geiger.
\newblock Multi-modal fusion transformer for end-to-end autonomous driving.
\newblock In \emph{CVPR}, 2021.

\bibitem[Qi et~al.(2017)Qi, Su, Mo, and Guibas]{qi2017pointnet}
Charles~R Qi, Hao Su, Kaichun Mo, and Leonidas~J Guibas.
\newblock Pointnet: Deep learning on point sets for 3d classification and segmentation.
\newblock In \emph{Proceedings of the IEEE conference on computer vision and pattern recognition}, pp.\  652--660, 2017.

\bibitem[Radford et~al.(2019)Radford, Wu, Child, Luan, Amodei, and Sutskever]{radford2019language}
Alec Radford, Jeff Wu, Rewon Child, David Luan, Dario Amodei, and Ilya Sutskever.
\newblock Language models are unsupervised multitask learners.
\newblock 2019.

\bibitem[Renz et~al.(2022)Renz, Chitta, Mercea, Koepke, Akata, and Geiger]{Renz2022CORL}
Katrin Renz, Kashyap Chitta, Otniel-Bogdan Mercea, A.~Sophia Koepke, Zeynep Akata, and Andreas Geiger.
\newblock Plant: explainable planning transformers via object-level representations.
\newblock In \emph{CoRL}, 2022.

\bibitem[Shao et~al.(2022)Shao, Wang, Chen, Li, and Liu]{shao2022interfuser}
Hao Shao, Letian Wang, RuoBing Chen, Hongsheng Li, and Yu~Liu.
\newblock Safety-enhanced autonomous driving using interpretable sensor fusion transformer.
\newblock \emph{CoRL}, 2022.

\bibitem[Shao et~al.(2023)Shao, Wang, Chen, Waslander, Li, and Liu]{shao2023reasonnet}
Hao Shao, Letian Wang, Ruobing Chen, Steven~L Waslander, Hongsheng Li, and Yu~Liu.
\newblock Reasonnet: End-to-end driving with temporal and global reasoning.
\newblock In \emph{Proceedings of the IEEE/CVF Conference on Computer Vision and Pattern Recognition}, pp.\  13723--13733, 2023.

\bibitem[Shi et~al.(2024)Shi, Jiang, Dai, and Schiele]{shi2024mtrmultiagentmotionprediction}
Shaoshuai Shi, Li~Jiang, Dengxin Dai, and Bernt Schiele.
\newblock Mtr++: Multi-agent motion prediction with symmetric scene modeling and guided intention querying, 2024.
\newblock URL \url{https://arxiv.org/abs/2306.17770}.

\bibitem[Su et~al.(2024)Su, Wu, and Yan]{su2024difsdegocentricfullysparse}
Haisheng Su, Wei Wu, and Junchi Yan.
\newblock Difsd: Ego-centric fully sparse paradigm with uncertainty denoising and iterative refinement for efficient end-to-end autonomous driving, 2024.
\newblock URL \url{https://arxiv.org/abs/2409.09777}.

\bibitem[Sun et~al.(2024)Sun, Lin, Shi, Zhang, Wu, and Zheng]{sun2024sparsedriveendtoendautonomousdriving}
Wenchao Sun, Xuewu Lin, Yining Shi, Chuang Zhang, Haoran Wu, and Sifa Zheng.
\newblock Sparsedrive: End-to-end autonomous driving via sparse scene representation, 2024.
\newblock URL \url{https://arxiv.org/abs/2405.19620}.

\bibitem[Vaswani et~al.(2017)Vaswani, Shazeer, Parmar, Uszkoreit, Jones, Gomez, Kaiser, and Polosukhin]{Vaswani2017AttentionIA}
Ashish Vaswani, Noam~M. Shazeer, Niki Parmar, Jakob Uszkoreit, Llion Jones, Aidan~N. Gomez, Lukasz Kaiser, and Illia Polosukhin.
\newblock Attention is all you need.
\newblock In \emph{NeurIPS}, 2017.
\newblock URL \url{https://api.semanticscholar.org/CorpusID:13756489}.

\bibitem[Vora et~al.(2020)Vora, Lang, Helou, and Beijbom]{vora2020pointpainting}
Sourabh Vora, Alex~H Lang, Bassam Helou, and Oscar Beijbom.
\newblock Pointpainting: sequential fusion for 3d object detection.
\newblock In \emph{CVPR}, pp.\  4604--4612, 2020.

\bibitem[Wang et~al.(2023)Wang, Liu, Wang, Li, and Zhang]{wang2023exploring}
Shihao Wang, Yingfei Liu, Tiancai Wang, Ying Li, and Xiangyu Zhang.
\newblock Exploring object-centric temporal modeling for efficient multi-view 3d object detection.
\newblock In \emph{Proceedings of the IEEE/CVF International Conference on Computer Vision}, pp.\  3621--3631, 2023.

\bibitem[Weng et~al.(2022)Weng, Ivanovic, Kitani, and Pavone]{weng2022whose}
Xinshuo Weng, Boris Ivanovic, Kris Kitani, and Marco Pavone.
\newblock Whose track is it anyway? improving robustness to tracking errors with affinity-based trajectory prediction.
\newblock In \emph{Proceedings of the IEEE/CVF Conference on Computer Vision and Pattern Recognition}, pp.\  6573--6582, 2022.

\bibitem[Weng et~al.(2024)Weng, Ivanovic, Wang, Wang, and Pavone]{Weng2024paradrive}
Xinshuo Weng, Boris Ivanovic, Yan Wang, Yue Wang, and Marco Pavone.
\newblock Para-drive: Parallelized architecture for real-time autonomous driving.
\newblock In \emph{Proceedings of the IEEE/CVF Conference on Computer Vision and Pattern Recognition (CVPR)}, 2024.

\bibitem[Wu et~al.(2022)Wu, Jia, Chen, Yan, Li, and Qiao]{wu2022trajectoryguided}
Penghao Wu, Xiaosong Jia, Li~Chen, Junchi Yan, Hongyang Li, and Yu~Qiao.
\newblock Trajectory-guided control prediction for end-to-end autonomous driving: A simple yet strong baseline.
\newblock In \emph{NeurIPS}, 2022.

\bibitem[Xie et~al.(2023)Xie, Kong, Zhang, Ren, Pan, Chen, and Liu]{xie2023robobev}
Shaoyuan Xie, Lingdong Kong, Wenwei Zhang, Jiawei Ren, Liang Pan, Kai Chen, and Ziwei Liu.
\newblock Robobev: Towards robust bird's eye view perception under corruptions.
\newblock \emph{arXiv preprint arXiv:2304.06719}, 2023.

\bibitem[Yang et~al.(2023{\natexlab{a}})Yang, Chen, Tian, Tao, Zhu, Zhang, Huang, Li, Qiao, Lu, et~al.]{yang2023bevformer}
Chenyu Yang, Yuntao Chen, Hao Tian, Chenxin Tao, Xizhou Zhu, Zhaoxiang Zhang, Gao Huang, Hongyang Li, Yu~Qiao, Lewei Lu, et~al.
\newblock Bevformer v2: Adapting modern image backbones to bird's-eye-view recognition via perspective supervision.
\newblock In \emph{Proceedings of the IEEE/CVF Conference on Computer Vision and Pattern Recognition}, pp.\  17830--17839, 2023{\natexlab{a}}.

\bibitem[Yang et~al.(2023{\natexlab{b}})Yang, Jia, Li, and Yan]{Yang2023LLM4DriveAS}
Zhenjie Yang, Xiaosong Jia, Hongyang Li, and Junchi Yan.
\newblock Llm4drive: A survey of large language models for autonomous driving.
\newblock \emph{ArXiv}, abs/2311.01043, 2023{\natexlab{b}}.

\bibitem[Zeng et~al.(2022)Zeng, Dong, Zhang, Wang, Zhang, and Wei]{zeng2022motr}
Fangao Zeng, Bin Dong, Yuang Zhang, Tiancai Wang, Xiangyu Zhang, and Yichen Wei.
\newblock Motr: End-to-end multiple-object tracking with transformer.
\newblock In \emph{European Conference on Computer Vision}, pp.\  659--675. Springer, 2022.

\bibitem[Zeng et~al.(2019)Zeng, Luo, Suo, Sadat, Yang, Casas, and Urtasun]{zeng2019nmp}
Wenyuan Zeng, Wenjie Luo, Simon Suo, Abbas Sadat, Bin Yang, Sergio Casas, and Raquel Urtasun.
\newblock End-to-end interpretable neural motion planner.
\newblock In \emph{CVPR}, 2019.

\bibitem[Zhai et~al.(2023)Zhai, Feng, Du, Mao, Liu, Tan, Zhang, Ye, and Wang]{zhai2023ADMLP}
Jiang-Tian Zhai, Ze~Feng, Jihao Du, Yongqiang Mao, Jiang-Jiang Liu, Zichang Tan, Yifu Zhang, Xiaoqing Ye, and Jingdong Wang.
\newblock Rethinking the open-loop evaluation of end-to-end autonomous driving in nuscenes.
\newblock \emph{arXiv preprint arXiv:2305.10430}, 2023.

\bibitem[Zhang et~al.(2024)Zhang, Wang, Zhu, Zhao, Chen, Zhang, Gong, Zhou, Zhang, Wang, Tan, Zhou, Xu, Yao, Zhang, Liu, Di, and Li]{zhang2024sparseadsparsequerycentricparadigm}
Diankun Zhang, Guoan Wang, Runwen Zhu, Jianbo Zhao, Xiwu Chen, Siyu Zhang, Jiahao Gong, Qibin Zhou, Wenyuan Zhang, Ningzi Wang, Feiyang Tan, Hangning Zhou, Ziyao Xu, Haotian Yao, Chi Zhang, Xiaojun Liu, Xiaoguang Di, and Bin Li.
\newblock Sparsead: Sparse query-centric paradigm for efficient end-to-end autonomous driving, 2024.
\newblock URL \url{https://arxiv.org/abs/2404.06892}.

\bibitem[Zhang et~al.(2022)Zhang, Tang, Geng, Chen, Xin, and Wang]{zhang2022mmfn}
Qingwen Zhang, Mingkai Tang, Ruoyu Geng, Feiyi Chen, Ren Xin, and Lujia Wang.
\newblock Mmfn: multi-modal-fusion-net for end-to-end driving.
\newblock \emph{IROS}, 2022.

\bibitem[Zhang et~al.(2021)Zhang, Liniger, Dai, Yu, and Van~Gool]{zhang2021roach}
Zhejun Zhang, Alexander Liniger, Dengxin Dai, Fisher Yu, and Luc Van~Gool.
\newblock End-to-end urban driving by imitating a reinforcement learning coach.
\newblock In \emph{ICCV}, 2021.

\end{thebibliography}
\bibliographystyle{iclr2025_conference}

\appendix
\section{Implementation Details}\label{sec:detail}
We implement the model with Pytorch. When comparing with state-of-the-art works, we report results of DriveTransformer-Large. When conducting ablation studies, we report results of DriveTransformer-Base on Dev10 benchmark if not specified. All models are trained in Bench2Drive~\citep{jia2024bench} base set (1000 clips) for 30 epochs on 8*A800 with a learning rate 1e-4, weight decay 0.05, dropout 0.1, AdamW, and cosine annealing schedule. We use ResNet50 as image backbones and the image size of (384, 1056) The temporal length ($T_{\text{queue}}$) are set as 10 (1 second 10Hz) in Bench2Drive and 4 (2 second 2Hz) in nuScenes. At each time-step, the Top-K queries is pushed into queue, where we set K as 50.  The initial number of queries is set as 900 for agent queries following~\citep{wang2023exploring} and 100 for map queries following~\citep{jiang2023vad}. \textbf{We will open source our code and checkpoints}. 

\section{Dev10 Benchmark}\label{sec:dev10}
Bench2Drive~\citep{jia2024bench} is a closed-loop evaluation protocol with 220 routes. The abundant number of routes could lower the variance of evaluation while introducing significant computational challenges. For example, it could take 2-3 days to evaluate DriveTransformer-Large on 8*A800. \rebcolor{To this end, for fast development of ideas, we propose \textbf{Dev10} Benchmark by the following procedure:}

1. \rebcolor{There are 44 kinds of scenarios in Benc2Drive where each type has 5 different routes under different locations and weathers. Due to the low variance of Bench2Drive's short routes, selecting one route per scenario could reflect the model's ability in that case.}

2. \rebcolor{Further, there are some very similar scenarios in these 44 types. For example, as shown in page 15 of Bench2Drive's paper (\url{https://arxiv.org/pdf/2406.03877}), the difference between scnenario 2 "ParkingCutIn", scenario 3 "ParkingCutIn", and scenario 4 "StaticCutIn" all examine the ability of the ego vehicle to slow down or brake for the cut-in vehicle. Thus, after discussing with the Bench2Drive official team, these scenarios could be summarized into 10 high-level types:}
\begin{itemize}
    \item \textbf{ParkingExit}: requiring the model to drive out of a parking lot.
\item \textbf{ParkingCrossingPedestrian}, DynamicObjectCrossing, ControlLoss, PedestrainCrossing, VehicleTurningRoutePedestrian, VehicleTurningRoute, HardBrake, OppositeVehicleRunningRedLight, OppositeVehicleTakingPriority: requiring the model to conduct emergency brake or slow down drasticly under dangerous situations.
\item \textbf{StaticCutIn}, HighwayExit, InvadingTurn, ParkingCutIn, HighwayCutIn: requiring the model to handle cut-in behaviors of the front vehicle*
\item \textbf{HazardAtSideLane}, ParkedObstacle, Construction, Accident: requiring the model to overtake the blocking obstacles  in front of it.
\item \textbf{YieldToEmergencyVehicle}: requiring the model to give  way to emergency vehicles.
\item \textbf{ConstructionObstacleTwoWays}, ParkedObstacleTwoWays, AccidentTwoWays, VehiclesDooropenTwoWays, HazardAtSideLaneTwoWays: requiring the model to drive in the reverse lane for a short distance, complete the overtaking, and then return to the original lane.
\item \textbf{NonSignalizedJunctionLeftTurn}, SignalizedJunctionLeftTurn, InterurbanActorFlow, InterurbanAdvancedActorFlow, CrossingBicycleFlow, VinillaNonSignalizedTurn, VinillaNonSignalizedTurnEncounterStopsign, VinillaSignalizedTurnEncounterGreenLight, VinillaSignalizedTurnEncounterRedLight, TJunction: requiring the model to handle the traffic at intersections and complete its turn.
\item \textbf{BlockedIntersection}: requiring the model to yield for the blocking event within the intersection until the event is finished.
\item \textbf{SequentialLaneChange}: requiring the model to continuously change several lanes.
\item \textbf{SignalizedJunctionLeftTurnEnterFlow}, EnterActorFlows, SignalizedJunctionRightTurn, NonSignalizedJunctionRightTurn, MergerIntoSlowTraffic, MergerIntoSlowTrafficV2: requiring the model to enter the dense traffic flow and merge into it.
\end{itemize}

3. \rebcolor{For the 10 high-level types, we select one route for each with diverse weathers and towns. 
We give the details of Dev10 below}:
\begin{table}[!h]
\centering
\caption{\textbf{Routes of Dev10 protocol.\vspace{-2mm}}}
\begin{tabular}{llll}
\toprule
Scenario                            & Route-ID & Road-ID & Town \\ \midrule
ParkingExit                         & 3514     & 892     & 13   \\ 
ParkingCrossingPedestrian           & 3255     & 1237    & 13   \\ 
StaticCutIn                         & 26405    & 137     & 15   \\ 
HazardAtSideLane                    & 25381    & 37      & 05   \\ 
YieldToEmergencyVehicle             & 25378    & -       & 03   \\ 
ConstructionObstacleTwoWays         & 25424    & 269     & 11   \\ 
NonSignalizedJunctionLeftTurn       & 2091     & -       & 12   \\ 
BlockedIntersection                 & 27494    & 16      & 04   \\ 
SequentialLaneChange                & 16569    & 1157    & 12   \\ 
SignalizedJunctionLeftTurnEnterFlow & 28198    & 234     & 15   \\ \bottomrule
\end{tabular}
\vspace{-3mm}
\end{table}

4. \rebcolor{We verify the variance of Dev10 with DriveTransformer-Base with 3 different seeds. The driving scores are 60.45, 59.20, 58.99 and the success rates 0.3, 0.3, 0.3, demonstrating very low variance}.

5. \rebcolor{When comparing with other methods, we stick to 220 routes to ensure fairness.  When conducting ablation studies, we use Dev10 to save computational resource}. 

6. \rebcolor{When comparing with other methods, we stick to 220 routes while conducting ablation studies, we use Dev10. Additionally, Dev10 could also serve as a validation set to avoid researchers overfitting Bench2Drive220.
It could also avoid the overfit of Bench2Drive220 so that it could be a test set. We will open source the Dev10 benchmark and the Bench2Drive official team plans to integrate it into their official repo to provide a short and economic validation set.}

\textbf{We will open source Dev10 protocol}.

\section{\rebcolor{Limitations}}
\rebcolor{Similar to existing end-to-end autonomous driving systems, DriveTrasnformer entangles the update of all sub-tasks and thus brings challenges for the maintenance of the whole system. An important future direction would make them less coupled and thus easier to debug and main separately.}

\begin{table}[!h]
\caption{\rebcolor{\textbf{Comparison of Performance on Middle Tasks.} $\dagger$ ParaDrive's latency is calculated by their claim that \textit{2.77x speed up compared to UniAD} when disabling all middle tasks.} \label{tab:sub-task}}
\centering
\resizebox{1.0\textwidth}{!}{
\begin{tabular}{l|cc|ccc|cc|c}
\toprule
\multicolumn{1}{c|}{\multirow{2}{*}{\textbf{Method}}} & \multicolumn{2}{c|}{\textbf{Detection}}          & \multicolumn{3}{c|}{\textbf{Motion}}                                                  & \multicolumn{2}{c|}{\textbf{Online Mapping}}               & \multirow{2}{*}{\textbf{Latency}} \\ \cline{2-8}
\multicolumn{1}{c|}{}                                 & \multicolumn{1}{c}{\textbf{NDS}} & \textbf{mAP} & \multicolumn{1}{c}{\textbf{minADE}} & \multicolumn{1}{c}{\textbf{minFDE}} & \textbf{MR} & \multicolumn{1}{c}{\textbf{IoU-Road}} & \textbf{IoU-Lane} &                                   \\ \hline
UniAD~\cite{hu2023planning}                                                  & \multicolumn{1}{c}{49.8}         & 38.0         & \multicolumn{1}{c}{0.72}            & \multicolumn{1}{c}{1.05}            & 0.15        & \multicolumn{1}{c}{0.30}              & 0.67              & 663.4ms                           \\ \hline
ParaDrive~\cite{Weng2024paradrive}                                              & \multicolumn{1}{c}{48.0}         & 37.0         & \multicolumn{1}{c}{0.72}            & \multicolumn{1}{c}{-}               & -           & \multicolumn{1}{c}{0.33}              & 0.71              & 239.5ms$\dagger$                           \\ \hline
\textbf{DriveTransformer}                              & \multicolumn{1}{c}{\textbf{59.3}} & \textbf{49.9} & \multicolumn{1}{c}{\textbf{0.61}}   & \multicolumn{1}{c}{\textbf{0.95}}   & \textbf{0.13} & \multicolumn{1}{c}{\textbf{0.39}}     & \textbf{0.77}     & \textbf{211.7ms}               \\ \bottomrule
\end{tabular}}
\end{table}

\section{\rebcolor{Comparison of Middle Tasks}}
\rebcolor{
We compare the performance of middle tasks in nuScenes validation set as in Table~\ref{tab:sub-task}.}

\section{\rebcolor{Comparison with Concurrent Parallel and Sparse based Methods}}
\rebcolor{Concurrent to DriveTransformer, there are other sparse-based methods including SparseAD~\cite{zhang2024sparseadsparsequerycentricparadigm} and SparseDrive~\cite{sun2024sparsedriveendtoendautonomousdriving}. To provide a comprehensive overview of existing sparse based methods, we compare DriveTransformer with them together with ParaDrive~\cite{Weng2024paradrive}, a most recent efficient modular E2E-AD method}:

\begin{table}[!ht]
\centering
\begin{tabular}{l|cccc|cccc|c}
\toprule
\multicolumn{1}{c|}{\multirow{2}{*}{\textbf{Method}}} & \multicolumn{4}{c|}{\textbf{L2 (m)}}                                                     & \multicolumn{4}{c|}{\textbf{Collision (\%)}}                                             & \multirow{2}{*}{\textbf{Latency}} \\ \cline{2-9}
\multicolumn{1}{c|}{}                                 & \multicolumn{1}{c}{1s}   & \multicolumn{1}{c}{2s}   & \multicolumn{1}{c}{3s}   & Avg. & \multicolumn{1}{c}{1s}   & \multicolumn{1}{c}{2s}   & \multicolumn{1}{c}{3s}   & Avg. &                                   \\ \midrule
ParaDrive                                              & \multicolumn{1}{c}{0.25} & \multicolumn{1}{c}{0.46} & \multicolumn{1}{c}{0.74} & 0.48 & \multicolumn{1}{c}{0.14} & \multicolumn{1}{c}{0.23} & \multicolumn{1}{c}{0.39} & 0.25 & 239.5ms                           \\ \midrule
SparseAD-B                                             & \multicolumn{1}{c}{0.15} & \multicolumn{1}{c}{0.31} & \multicolumn{1}{c}{0.57} & 0.35 & \multicolumn{1}{c}{0.00} & \multicolumn{1}{c}{0.06} & \multicolumn{1}{c}{0.21} & 0.09 & 285.7ms                           \\ 
SparseAD-L                                             & \multicolumn{1}{c}{0.15} & \multicolumn{1}{c}{0.31} & \multicolumn{1}{c}{0.56} & 0.34 & \multicolumn{1}{c}{0.00} & \multicolumn{1}{c}{0.04} & \multicolumn{1}{c}{0.15} & 0.06 & 1428.6ms                          \\ \midrule
SparseDrive-S                                          & \multicolumn{1}{c}{0.29} & \multicolumn{1}{c}{0.58} & \multicolumn{1}{c}{0.96} & 0.61 & \multicolumn{1}{c}{0.01} & \multicolumn{1}{c}{0.05} & \multicolumn{1}{c}{0.18} & 0.08 & 111.1ms                           \\
SparseDrive-B                                          & \multicolumn{1}{c}{0.29} & \multicolumn{1}{c}{0.55} & \multicolumn{1}{c}{0.91} & 0.58 & \multicolumn{1}{c}{0.01} & \multicolumn{1}{c}{0.02} & \multicolumn{1}{c}{0.13} & 0.06 & 137.0ms                           \\ \midrule
DriveTransformer-S                                     & \multicolumn{1}{c}{0.19} & \multicolumn{1}{c}{0.33} & \multicolumn{1}{c}{0.66} & 0.39 & \multicolumn{1}{c}{0.01} & \multicolumn{1}{c}{0.07} & \multicolumn{1}{c}{0.21} & 0.10 & 93.8ms                            \\
DriveTransformer-B                                     & \multicolumn{1}{c}{0.16} & \multicolumn{1}{c}{0.31} & \multicolumn{1}{c}{0.56} & 0.34 & \multicolumn{1}{c}{0.01} & \multicolumn{1}{c}{0.06} & \multicolumn{1}{c}{0.16} & 0.08 & 139.6ms                           \\ 
DriveTransformer-L                                     & \multicolumn{1}{c}{0.16} & \multicolumn{1}{c}{0.30} & \multicolumn{1}{c}{0.55} & 0.33 & \multicolumn{1}{c}{0.01} & \multicolumn{1}{c}{0.06} & \multicolumn{1}{c}{0.15} & 0.07 & 221.7ms                           \\ \bottomrule
\end{tabular}
\end{table}

\rebcolor{We could observe that DriveTransformer achieves good L2 with high efficiency}.

\section{\rebcolor{Training Stability \& Multi-Stage Training}}
\rebcolor{In pioneering work UniAD~\cite{hu2023planning}, they adopt a three step training strategy: (1) Training a BEVFormer~\cite{li2022bevformer} with 3D object detection task; (2) Training TrackFormer and MapFormer; (3) Training all modules together. They explain in their paper that "\textit{We first jointly train perception parts, i.e., the tracking and mapping modules, for a few epochs (6 in our experiments), and then train the model end-to-end for 20 epochs with all perception, prediction and planning modules. The two-stage training is found more stable empirically}." and "\textit{Joint learning. UniAD is trained in two stages which we find more stable.}".} 

\rebcolor{We conduct experiments to train UniAD with one single stage (loading pretrained BEVFormer) in nuScenes shows that the overall loss (around 54) stops decreasing in early epoch (around epoch 4) while UniAD trained with the official two stage training has the final loss of (around 34), as shwon in Table~\ref{tab:uniad-stage}}.

\begin{table}[!ht]
\centering
\caption{\rebcolor{\textbf{Comparison of One-Stage and Two-Stage Trainging for UniAD.}\label{tab:uniad-stage}}}
\resizebox{1.0\textwidth}{!}{
\begin{tabular}{l|cccccc}
\toprule
\textbf{Method} & \textbf{Final Loss} & \textbf{NDS (Detection)} & \textbf{IoU-lane (Mapping)} & \textbf{AMOTA(Tracking)} & \textbf{minADE(Motion)} & \textbf{Avg. L2 (Planning)} \\ \midrule
One-Stage       & 54.07               & 38.1                     & 23.5                        & 20.4                     & 2.52                    & 2.01                        \\ 
Two-Stage       & \textbf{34.21}      & \textbf{49.8}            & \textbf{30.2}               & \textbf{35.9}            & \textbf{0.71}           & \textbf{1.03}               \\ \bottomrule
\end{tabular}}
\end{table}
\rebcolor{We could observe that one-stages result in underfit of all modules with much higher final loss}.

\rebcolor{Further, in concurrent work SparseAD~\cite{zhang2024sparseadsparsequerycentricparadigm}, they mention that \textit{The training is divided into \textbf{three stages in total}...stage 1 and stage 2 can be merged during training for a shorter training time with slight performance degradation in exchange.}. In SparseDrive~\cite{sun2024sparsedriveendtoendautonomousdriving}, they mention that \textit{The training of SparseDrive is divided into \textbf{two stages}. In stage-1, we train symmetric sparse perception module from scratch to learn the sparse scene representation. In stage-2, sparse perception module and parallel motion planner are trained together.}}

\rebcolor{In contrast, compared to the sequential designs (UniAD, SparseAD, SparseDrive), one significant improvement in DriveTransformer is that \textbf{all tasks' interaction is learnt via attention instead of manual ordering}. As a result, at the early stage of training, each task could access information directly thorough sensor cross attention and temporal self attention, reducing the reliance on other under-trained tasks' queries. Such designs are friendly to the scaling up and industrial application of E2E-AD methods. As shown in Table~\ref{tab:paradigm}, \textit{Pretrain Perception} does not provide gains for DriveTransformer, which proves the training stability of DriveTransformer.}

\end{document}